# A reproducible approach to merging behavior analysis based on High Definition Map

Yang Li, Yang Liu, Daiheng Ni, Ang Ji, Linbo Li*, Yajie Zou

*Yang Li*
Ph.D., Candidate
Key Laboratory of Road and Traffic Engineering of Ministry of Education,
Tongji University, China,
Email: cc960719@tongji.edu.cn

*Yang Liu*
Ph.D., Assistant Professor
Department of Civil and Environmental Engineering and the Department of Industrial Systems Engineering and Management,
National University of Singapore, Singapore
Email: iseliuy@nus.edu.sg

*Daiheng Ni*
Ph.D., Professor
Civil and Environmental Engineering,
University of Massachusetts Amherst, Massachusetts 01003, USA
Email: ni@engin.umass.edu

*Ang Ji*
Ph.D., Assistant Professor
School of Transportation and Logistics,
Southwest Jiaotong University, Chengdu, China
Email: ang.ji@swjtu.edu.cn

*Linbo Li\* (Corresponding Author)*
Ph.D., Associate Professor
Key Laboratory of Road and Traffic Engineering of Ministry of Education,
Tongji University, China,
Email: llinbo@tongji.edu.cn

*Yajie Zou*
Ph.D., Associate Professor
Key Laboratory of Road and Traffic Engineering of Ministry of Education,
Tongji University, China,
Email: yajiezou@hotmail.com

1# A reproducible approach to merging behavior analysis based on High Definition Map

Yang Li, Yang Liu, Daiheng Ni, Ang Ji, Linbo Li, Yajie Zou*Abstract*—Existing research on merging behavior generally prioritize the application of various algorithms, but often overlooks the fine-grained process and analysis of trajectories. This leads to the neglect of surrounding vehicle matching, the opaqueness of indicators definition, and reproducible crisis. To address these gaps, this paper presents a reproducible approach to merging behavior analysis. Specifically, we outline the causes of subjectivity and irreproducibility in existing studies. Thereafter, we employ lanelet2 High Definition (HD) map to construct a reproducible framework, that minimizes subjectivities, defines standardized indicators, identifies alongside vehicles, and divides scenarios. A comparative macroscopic and microscopic analysis is subsequently conducted. More importantly, this paper adheres to the Reproducible Research concept, providing all the source codes and reproduction instructions. Our results demonstrate that although scenarios with alongside vehicles occur in less than 6% of cases, their characteristics are significantly different from others, and these scenarios are often accompanied by high risk. This paper refines the understanding of merging behavior, raises awareness of reproducible studies, and serves as a watershed moment.

## I. Introduction

### A. Why merging behavior analysis is important?

NUMEROUS studies have indicated that the advent of Autonomous Vehicles (AV) could improve traffic safety, and efficiency, as well as alleviate traffic congestion, and fuel consumption [1-3]. With the commercial mass production of L2-class vehicles, and the impending implementation of L4/L5 technologies, the application of autonomous driving technology is facing unprecedented challenges. Improving the interaction capability of AVs has become a crucial topic, requiring a deeper understanding of various driving scenarios than previous studies. This paper focuses on the merging behavior, which is one of the most important driving tasks. Merging behavior is a common occurrence on expressways, where vehicles move to the mainline lanes via on-ramps or acceleration lanes with suitable gaps.

Since this behavior involves strong interaction with surrounding vehicles, it often results in recurring bottlenecks [4], and traffic conflicts or even accidents [5]. The importance of merging behavior analysis could be summarized as follows: a) To provide reliable data pre-processing and analysis support for data-driven planning and controlling algorithms, enhancing the human-like interaction ability and prediction accuracy of AVs during the merging process; b) To guide the construction of cases in autonomous driving simulation test, and generalize the parameters of merging scenario more reasonably. It can also identify corner cases and improve the generalizability of AVs; c) To study the characteristic of different vehicles types, drivers, weather, and working days, which could help in the setting default parameters in Advanced Driver Assistance System. d) To guide the merging scenario setting of microscopic traffic simulation software for scenarios, such as SUMO and VISSIM.

### B. What has yet been done?

Despite that many studies have revealed drivers' merging behavior from different aspects over the years, there are still some issues remained to be addressed. First, existing studies have not considered fine-grained multi-vehicle interaction scenarios. They only focused on the matching of lead and rear vehicles, while neglecting the potential presence of alongside vehicles during the merging process. And the interaction distance threshold between two vehicles has not been studied explicitly. Second, existing studies are more of less subjective for the definition and analysis of indicators. At the same time, they tend to focus on micro-level effects without incorporating the influence of macro-level traffic flow. Third, a large number of existing studies may suffer a reproducibility crisis, and the obtained outcomes can be difficult to be replicated by others. Overall, there are still huge research gaps to be addressed. The lack of an in-depth analysis makes it difficult to grasp the nature of this behavior and makes present simulation software hard to depict this behavior accurately.

### C. What can be done better?

To address these gaps, we conduct a detailed investigation of merging behavior characteristics using High Definition (HD) map. The fuse of HD map data and merging trajectory allows us to extract trajectory, match vehicles, define indicators, classify

This manuscript is submitted on February 30, 2023 (Corresponding author: Linbo Li). This work is supported by National Key R&D Program of China (grant numbers 2019YFB1600703) and National Natural Science Foundation of China (grant numbers 52172331).

Yang Li, Linbo Li, Yajie Zou are with Key Laboratory of Road and Traffic Engineering of the Ministry of Education, Tongji University, 4800 Cao'an Road, Shanghai, P.R. China (e-mail: cc960719@tongji.edu.cn, llinbo@tongji.edu.cn, yajiezou@hotmail.com).

Yang Liu is with the Department of Civil and Environmental Engineering and the Department of Industrial Systems Engineering and Management at the National University of Singapore. (e-mail: iseliuy@nus.edu.sg).

Ang Ji is with School of Transportation and Logistics, Southwest Jiaotong University, Chengdu, China (e-mail: ang.ji@swjtu.edu.cn).

Daiheng Ni is with department of Civil and Environmental Engineering, the University of Massachusetts Amherst, Massachusetts 01003, USA (ni@engin.umass.edu). He is no involvement in the research grants.

scenarios, macroscopic and microscopic analysis more objectively and finely. In addition, to eliminate reproducibility crisis that may occur in this study, a transparent analysis is presented as shown in Fig. 1. While everyone is trying to apply various data-driven algorithms, should we be vigilant about whether their trajectory processing and analysis are reliable and replicable. We hope that this "unconventional" paper will assist us in better comprehending merging behavior, particularly for the large number of recent published studies. The remainder of this paper is organized as follows: Section II presents the related work, while Section III elaborates on reproducible framework. Sections IV and V present the analysis results. Section VI exhibits the findings and discussion. Finally, Section VII ends with the conclusion.

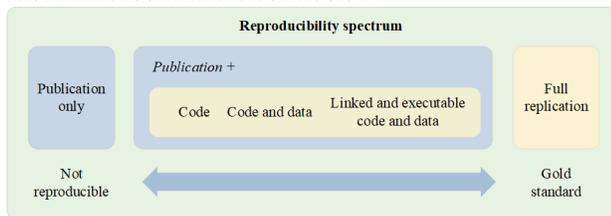

**Fig. 1 The spectrum of reproducibility [6]**

## II. RELATED WORK

Merging behavior has attracted a lot of research interest. This section reviews existing representative studies related to this paper.

### A. Existing merging behavior analysis

Up to now, the commonly-used traffic public datasets includes NGSIM [7], HighD [8], INTERACTION [9, 10], as well as some private dataset [5, 11, 12]. The widely-used methods involve classification, prediction, and regression models, etc. Li [13] reviewed the past 15 years of research using vehicle trajectory data. Di [14] systematically reviewed physics-based and AI-based studies in the mixed traffic flow environment, and they also reviewed several publicly autonomous driving datasets. Based on these datasets, scholars conduct a series of analysis from macroscopic and microscopic levels. Wan [15] proposed a novel sequential decision-making model for the candidate gap generation, targeting gap selection, merging location choice, and acceleration decisions using NGSIM dataset. Sarvi [16] conducted a comprehensive analysis of traffic behavior and characteristics during freeway ramp merging under congested traffic conditions. Wang [9] provided quantitative analysis and evaluation of the merging behavior at highway on-ramps with congested traffic in a volume of time and space. Zhang [10] introduced Hidden Marko Model into INTERACTION dataset, and obtained nine merging behavior patterns. Li [7] analyzed the influence of traffic density on merging position, speed, duration, and accepted gap based on NGSIM dataset. Gu [11] proposed a framework to explore the crash risk during the entire merging implementation. Daamen [12] investigated the merge location, accepted and offered gaps, and relaxation phenomenon.

### B. Reproducible Research

The field of science will not change overnight, but simply bringing the notion of reproducibility to the forefront and making it routine will make a difference [6]. Scholars from a variety of disciplines have emphasized the value of reproducible research, particularly in the fields of medicine [17] and artificial intelligence [18, 19]. As pointed by Peng [6], computational science has led to exciting new developments, but the nature of the work has exposed limitations in our ability to evaluate published findings. Recently, the significance of reproducible research has gained prominence in the field of transportation research [20, 21]. Zheng [20] considered reproducibility research is an unstoppable and inevitable future. Holland [21] pointed out "Without standardization researchers cannot compare "apples-to-apples," preventing us from knowing with certainty if a new technique represents a methodological advancement or if it simply benefits from a different interpretation of a given dataset." Siddiqui [22] presented a framework for establishing targets for national system performance measures for reliability. Szufel [23] proposed a multiagent, large-scale, vehicle routing modeling framework for the simulation of transportation system. Their model and the framework are all public on the Github website. Punzo [24] adopted the concept of reproducible research and introduced it to the car-following model calibration. Zheng [20] systematically introduced the importance, benefits, and challenges of reproducible research, and presented a template for doing reproducible research. Holland [21] examined irreproducibility that arises from the lack of standardization in network traffic analysis.

### C. Existing research gaps

However, the existing studies have following research gaps. a) Insufficient fine-grained division of merging scenarios: Current scenario division methods are frequently straight applications of various algorithms, with no thorough division in the extraction process. Existing studies tend to ignore the relative position changes between merging vehicles and their surroundings, especially for the alongside vehicles [4, 5, 7, 9-11]. However, the merging scenario with alongside vehicles might be more challenging for drivers, and ignoring these precious scenarios may lead to a partial understanding of merging behavior. b) Existence of a large amount of subjectivity: The process of trajectory extraction, vehicle matching, indicators definition, indicators calculation, and division of upstream/downstream merging areas is often accompanied by a strong subjectivity or is not explained in detail. For example, the determination of the starting and ending frame when extracting merging trajectory, the determination of the starting and ending point in space-time range during indicators definition, and the determination of the distance threshold when matching vehicles, for example how far away from itself the cars should be recognized as lead and rear vehicles. c) Irreproducible research: All existing merging studies may suffer a reproducible crisis. According to [25], nearly 70% of researchers have attempted and failed to duplicate another scientist's studies, and more than half are in the fields of engineering and physics. The top five main causes of irreproducible research include selective reporting, pressure to publish, a lack of statistical power or faulty analysis, insufficient replication in the original lab, and inadequate supervision or mentoring. As for merging behavior, the fine-tuning of certain parameters may result in large discrepancies in the results, affecting the credibility of these results. This inevitably leads to a reproducible crisis spreading in this field.

*D. The contributions of this work*

To address the above gaps, our contributions are given: a) Fine-grained alongside vehicle identification and scenario classification: we employ HD map for trajectory extraction, alongside vehicle identification, and scenario classification. Furthermore, the Jensen Shannon divergence is introduced to compare the similarities between our divided scenarios. b) Reduction of subjectivity throughout the analysis: we outline the causes of subjectivity and irreproducibility in exiting studies. Thereafter, we present methods to reduce subjectivity based on HD map, which include: construction of reproducible framework, definition of basic concepts of merging behavior, definition of standardized metrics. c) Construction of a reproducible analysis framework: We presents a fully transparent and reproducible analysis. All the source codes, project description, and reproduction guidance are provided. As pointed by [6], "Reproducibility has the potential to serve as a minimum standard for judging scientific claims when full independent replication of a study is not possible."

## III. ACHIEVING HIGH LEVEL OF DETAILS IN MERGING BEHAVIOR ANALYSIS

*A. Framework of high level of detail analysis*

Fig. 2 depicts our analysis framework. We provide a high level of details in merging behavior analysis from these perspectives: a) application of HD map; b) definition of standardized indicators; c) fine-grained alongside vehicle identification and scenario classification; d) reproducible microscopic analysis of merging point, distance, distance ratio, duration, lateral speed, lateral acceleration, safety analysis, two consecutive lane-change duration; e) reproducible macroscopic analysis of traffic flow, density, and speed; f) Provision of all code and intermediate processes. We employ exiD dataset, which was collected in 2021 [26]. ExiD dataset was gathered over the course of more than 16 hours at seven different locations along German highway exit/entry ramps, with a total of 69,172 vehicles. Locations 1, 2, and 4 record more than 10,000 vehicles, whereas locations 5 and 6 only record 5,000 to 6,000 vehicles. We focus on a single merging behavior, and adopts locations 2,3,5,6 for analysis as shown in Fig. 3. This is because that location 0 involves two consecutive merging behavior, location 1 is an intertwined area, and location 4 does not captured at the end of the acceleration lane.

The main advantages of exiD dataset compared to existing datasets can be summarized in two points, making it more suitable for merging behavior analysis among existing publicly datasets. a) exiD dataset contains more driving scenarios than NGSIM, HighD, Interaction-highways, and automatum datasets [26]. The collection position error is less than 10cm, and it provides three different types of vehicles: car, truck, and van. b) While none of the existing trajectory datasets provide HD map of the merging areas, exiD dataset provides Opendrive and lanelet2 HD maps, thereby helping us gain a deeper understanding of the merging behavior.

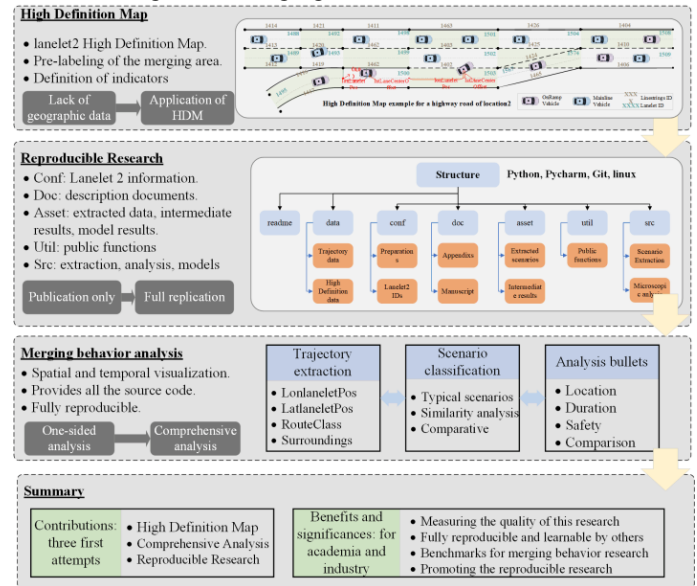

**Fig. 2 Our reproducible approach to merging behavior**

*B. Causes of subjectivity and irreproducibility*

*1) Difference in definition of basic concepts*

Since the merging process takes place over a period of time, it involves the determination of the starting and ending point., which is often accompanied by subjectivity. The existing definition of the starting point includes: a) the moment when the vehicle first gets on the ramp; b) the moment when the vehicle first enters the acceleration lane from the on ramp; c) the moment when the vehicle is ready to turn the wheel and begin to merge after driving in the acceleration lane for a period of time. The existing definition of the ending point includes: a) the moment when the vehicle enters the mainline; b) The center point of vehicle is located on the lane line between the acceleration lane and the main lane; c) the moment when the vehicle reaches the centerline of the mainline lane; d) the moment when the vehicle has been driving steadily in the mainline lane for a period of time. Overall, the merging process has at least $3 \times 4 = 12$ definitions. The third definition of the starting point, as well as the third and fourth definitions of the ending point are the most subjective. This is because: a) it is challenging to decide which frame should be used as starting or ending point for slow-steering vehicles; b) some drivers require more time to adjust their lateral position after entering mainline lane; c) different drivers and vehicle types may have different preferences for the selection of the points.

*2) Difference due to datasets, language, quality and habits*

Differences in the choice of locations and time periods in public datasets may also lead to large differences in results. For personal reasons or regulatory confidentiality, the results obtained from non-public datasets are difficult for others to reproduce. The programming language vary from different researchers. The most commonly used are Python, Matlab, R, etc. The results made by researchers with programming experience are more likely to be reproduced by others.

*3) Difference in indicators definition and calculation*

Existing studies tend to be more or less subjective in time-range and space-range of indicators definition. When defining indicators such as speed, acceleration, and safety, it is important to clarify whether we are looking at a specific point in time or at a statistical average over a period of time. The analysis of traffic flow, density and speed needs to define the spatial range.

*4) Difference in trajectory extraction and vehicle matching*

The parameter settings and judgment criterion of trajectory extraction in existing studies are a black box. Existing studies do not discuss the setting of distance thresholds when performing surrounding vehicle matching. These processes are susceptible to some degree of subjectivity.

*5) Difference in algorithm or visualization parameters*

The fine-tuning of some parameters in the algorithm and visualization process may lead to different results, such as the prediction algorithm, cluster algorithm, the maximum display scale of the horizontal axes, and the parameters of the boxplot to display outliers.

*C. Methods to minimize subjectivity and irreproducibility*

We reduce the subjectivity and irreproducibility of this study by the following points:

*1) Clarification of the basic concepts of merging behavior*

We define the basic concepts of merging behavior, such as the starting and ending points of the entire merging process. This is explained in detail in subsection D of this chapter.

*2) Clarification of the time-space range of each indicator*

We define standardized indicators. This is explained in detail in subsection E of this chapter.

*3) Clarification of the merging trajectory extraction process*

We identify the position of each vehicle at each frame more finely and unify the judgment criteria in trajectory extraction. Only three parameters are involved in the extraction process. This is explained in detail in subsection F of this chapter.

*4) Clarification of the setting of the distance threshold*

We set the distance threshold and explored whether different distance thresholds would affect the results. This is explained in detail in subsection B of the next chapter.

*5) Clarification of the identification of nearby vehicle*

We identify the lead, rear, and alongside vehicles that exist throughout the merging process, instead of only identifying surrounding vehicles at a certain moment. This is explained in detail in Section B of the next chapter.

*6) Clarification of the scenario classification*

We divide the scenarios according to the relative position changes of nearby vehicles. This is explained in detail in Section B of the next chapter.

*7) Full transparent and reproducible analysis*

All the analysis and results in this study are publicly availale on Github. Any of our analysis processes are transparent and reproducible.

*D. Application of Lanelet2 HD map*

*1) Lanelet2 format*

HD map in autonomous driving technology contains details which are not normally available in conventional maps. Conventional map primarily assists drivers with navigation, and its coordinate accuracy is about 10 meters. However, autonomous vehicles need to precisely determine their position and sense the surroundings objects to avoid any potential hazards; the absolute accuracy must be maintained within 1 meter. exiD dataset provided lanelet2 HD map. Lanelet2 format contains elements such as point, linestrings, polygons, lanelets, areas, and regElem. Points, which are the most fundamental building blocks and consist of distinctive IDs, 3D coordinates, and attributes; Area, represents the area that cannot be moved. Linestring, describes the shape of a map element by using two or more points. Laneletwidth identifies the atomic region of a map lane where directional movement occurs. RegElem is a general term for expressing traffic rules.

*2) HD map data pre-labeling*

We take location 2 as an example to illustrate the application of lanelet2 HD map. Fig. 4 shows the lanlet2 map for location 2. We divide the merging area of location2 into: Area 1: Vehicles approach the acceleration lane from the on-ramp. Since the left lane line is solid, vehicles cannot drive into the mainline lane in this area. The current laneletId is 1500, with a total length of 67.56m. Area 2: The left side of a vehicle is a dashed line, and it could merge into the mainline within this area. The current laneletId is 1503, with a total length of 119.57m. Area 3: The acceleration lane gradually disappears and lane drops from three to two. The current laneletId and length are 1567 and 40.65m. Area 4: This area is the upstream of the mainline outer lane within the merging area. It contains three laneletIds, 1489, 1493 and 1499, with lengths of 81.7m, 29.78m, and 67.87m. Area 5: vehicles travel from the acceleration lane to this area and may interact with mainline vehicles. The current laneletId is 1502, 1574 and the length is 119.45m, 40.65m. Similarly, we divide locations 3, 5, 6 into five areas with laneletId and length as shown in Tab. 1.

**Tab. 1 The laneletId and length for four locations**

| Area | Variable | Location | | | |
|---|---|---|---|---|---|
| | | 2 | 3 | 5 | 6 |
| 1 | laneletId | 1500 | 1415 | 1409 | 1459 |
| | length | 67.56 | 17.9 | 66.54 | 26.91 |
| 2 | laneletId | 1503 | 1525 | 1412 | 1514 |
| | length | 119.67 | 167.54 | 132.63 | 192.32 |
| 3 | laneletId | 1567 | 1531 | 1483 | 1513 |
| | length | 40.65 | 32.87 | 42.21 | 27.71 |
| 4 | laneletId | 1489/ 1493/ 1499 | 1405/ 1412/ 1414 | 1447/ 1450/ 1408 | 1459/ 1454 |
| | length | 81.72/ 29.78/ 67.87 | 90.78/ 25.86/ 17.9 | 31.9/ 57.93/ 66.05 | 26.91/ 43.13 |
| 5 | laneletId | 1502/ 1574 | 1524/ 1528 | 1411/ 1414 | 1463/ 1467 |
| | length | 119.45/ 40.65 | 168.03/ 32.49 | 134.39/ 42.19 | 191.65/ 27.4 |

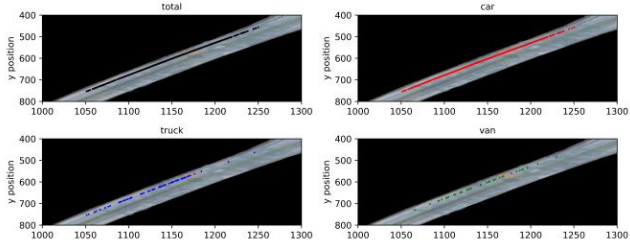

a) Location 2

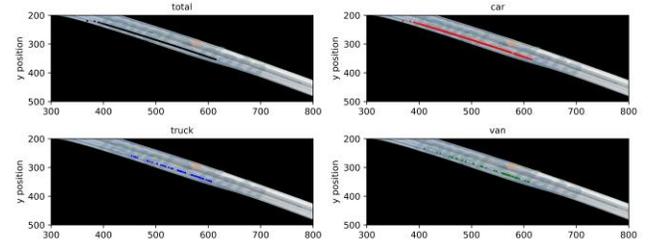

b) Location 3

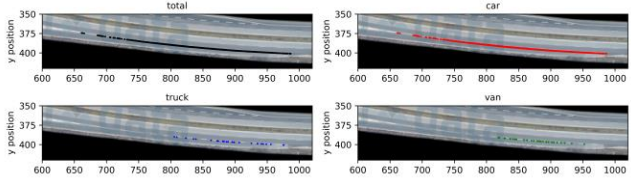

c) Location 5

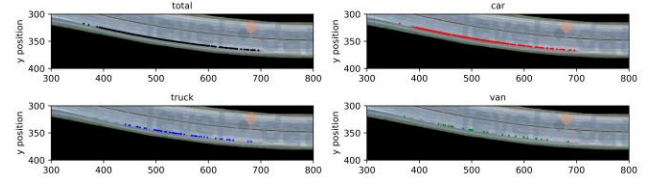

d) Location 6

**Fig. 3 The scatter plot of merging points of different types of vehicles under different locations**

The lower part of Fig. 4 shows an entire trajectory, where a vehicle goes from position A to position H. Position A (laneletId = 1495): the vehicle is in the on-ramp area. Position B (laneletId = 1500): current position is the junction of the on-ramp and the acceleration lane. Position C (laneletId = 1500): current road is an acceleration lane, and the left side is a solid line. Position D (laneletId = 1503): the dividing point between the solid line and the dashed line of the acceleration lane. Position E (laneletId = 1503): due to the left dashed line, the vehicle could merge from here. Position F (laneletId = 1502): the centerline of the vehicle is located at the junction of the acceleration lane and the mainline lane. The lanletoffset value at this moment changed from 2 jumps to -2. Position G (laneletId = 1574): the whole body of the vehicle merged into the mainline. Position H (laneletId = 1504): the current vehicle changed from the outer lane to the inner lane.

*3) Advantages of fusing HD map into trajectory analysis*

The advantages of incorporating HD map data into merging trajectory ("1+1>2") can be summarized as: a) More fine-grained recognition of each frame of trajectory: The application of HD map could identify the spatial-temporal location of each vehicle at each frame, such as the frame IDs of position B, D and F as shown in Fig. 4; b) More fine-grained alongside vehicle identification and scenario classification: The deeper recognition of the trajectory at each frame allows us to investigate the setting of distance thresholds, vehicle matching, and scenario classification. c) More objective indicators definition: each trajectory is subdivided according to Fig. 4. This allows us to unify the judgment criteria, thus making the definition of indicators more objective. d) Support for reproducible research: providing a solid support for our reproducible research.

*E. Definition of standardized indicators*

We define standardized indicators with the help of lanelet2 HD map.

a) Merging point: Each point represents the frame where the center point of the vehicle changes from the current lane to the mainline lane. That is, the latLaneCenterOffset values changes from 2 to -2. This corresponds to position F in Fig. 4.

b) Merging distance: The distance between the frame where the center point of the vehicle changes from the acceleration lane to the mainline lane. This corresponds to the distance between position D and position F. This indicator could reflect the preference for the merging location.

c) Merging distance ratio: Since the length of acceleration lanes varies from location to location, it is necessary to standardize the merging point characteristics. Thus, the merging distance of each vehicle is divided by the total length at each location, thus scaling the values of all vehicles to 0~1.

d) Merging duration: There are two definitions of duration in the existing studies. One is the duration from position E to F. It represents the interval time from the center of the vehicle, traveling from the center of acceleration lane to the junction of the mainline and the acceleration lanes. The other is from position D to position F. It represents the duration from the moment when the vehicle is ready for merge to the moment when the center of the vehicle has just entered the mainline lane. The second indicator is chosen since the starting point in the first is difficult to be discerned.

$$MergingDistance = x_F - x_D \quad (1)$$

$$MergingDistanceRatio = \frac{MergingDistance}{length(Area2 + Area3)} \quad (2)$$

$$MergingDuration = t_F - t_D \quad (3)$$

Where $x_F, x_D, t_F, t_D$ represents the longitudinal location and time of position F and D, $length(Area2 + Area3)$ represents the total length of Area 2 and Area 3.

e) Lateral speed and acceleration: The lateral speed and acceleration from the acceleration lane into mainline lane can also characterize drivers' preferences for lateral merging behavior. Therefore, we choose the time range from position D to F. These values can be obtained from latLaneCenterOffset field using lanelet2 HD map. We employ a fifth order polynomial to fit the discrete latLaneCenterOffset values to obtain smooth speed and acceleration curves.

f) Continuous lane-changing duration: A vehicle that has just merged into the mainline lane may immediately change lanes again to the innermost lane. For example, the vehicle in Fig. 4 changes lanes from position F to position H.

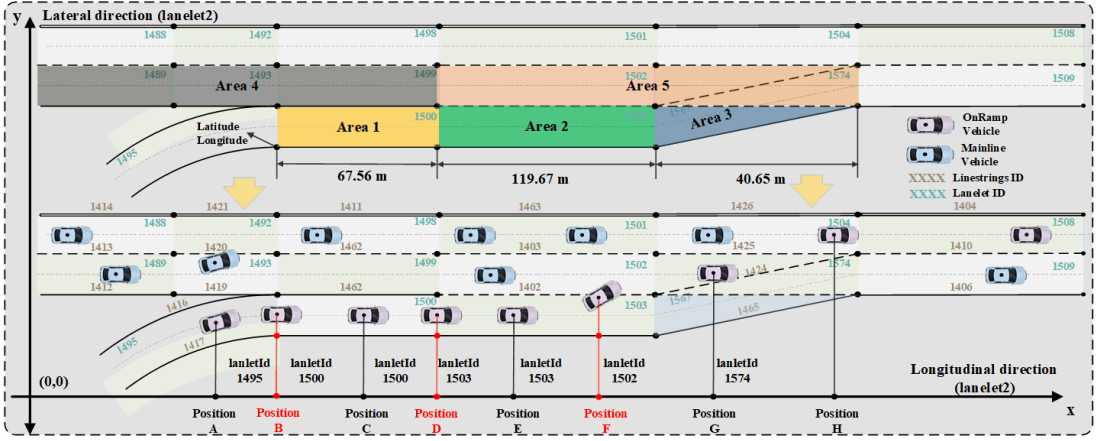

Fig. 4 The lanelet2 HD map for the merging area of Location 2

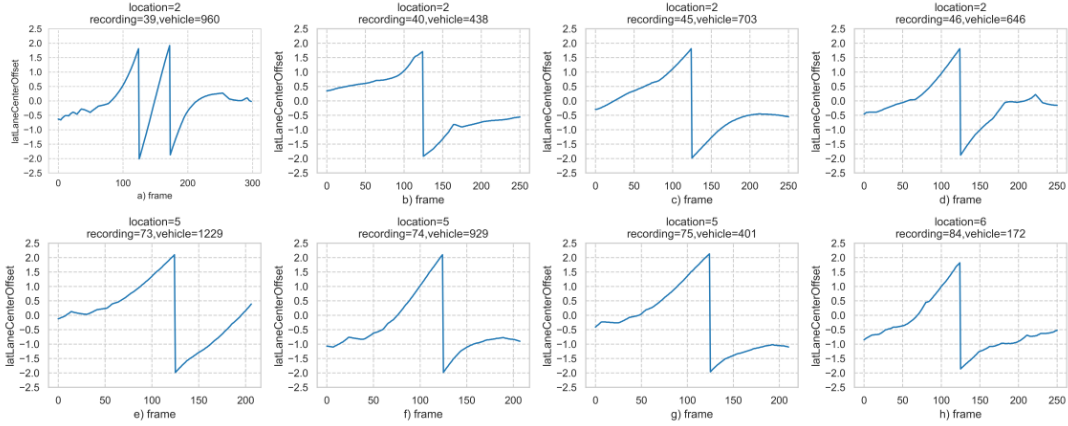

Fig. 5 Extracted merging trajectory examples

g) Two-dimensional time to collision (TTC): The traditional TTC could only calculate the car-following risk on the same lane, and is not suitable for vector calculation in two-dimensional space. Therefore, we adopt the vector calculation method of TTC in two-dimensional space [27]. This method does not need to distinguish the lane where the vehicle is located, and only needs to represent the vehicle position and velocity in the equation of motion as two-dimensional vector information, and only vector operations are required in the calculation process. We choose the time range of the vehicle from position D to position F. The minimum TTC with the lead and rear vehicles in that time range is selected. The calculation formulas are:

$$TTC(t) = \frac{x_i(t) - x_j(t) - l_i}{v_i(t) - v_j(t)} \quad (4)$$

$$d_{ij} + \dot{d}_{ij} TTC = 0 \quad (5)$$

$$d_{ij} = \sqrt{(p_i - p_j)^T (p_i - p_j)} \quad (6)$$

$$\dot{d}_{ij} d_{ij} = (p_i - p_j)^T (v_i - v_j) \quad (7)$$

Where $d_{ij}$ is the relative distance between the two vehicles, $\dot{d}_{ij}$ is the vector expression of the relative speed of the two vehicles approaching. $p_i$ and $p_j$ are the position coordinate vectors of the front and rear cars respectively. $v_i$ and $v_j$ are the speed vectors of the front and rear cars respectively.

$$ContinuousLaneChangeDuration = t_H - t_F \quad (8)$$

h) Traffic flow, traffic density, and space-mean speed: traffic flow characteristics may also be vital variables that affect merging behavior. Therefore, we selected two areas for comparative analysis, one is upstream area4, and the other is area5. We choose the time range from position B to F since these variables may influence the driver's decision behavior at the beginning moment of position B. We adopt the fundamental approach to estimate flow, density and speed. For the method, please refer to [28, 29]. The formulas are as follows:

$$|A| = L \times T \quad (9)$$

$$d(A) = \sum_{i=1}^{N} \Delta x_i \quad (10)$$

$$t(A) = \sum_{i=1}^{N} \Delta t_i \quad (11)$$

$$q_{flow} = d(A)/|A| \quad (12)$$

$$k_{density} = t(A)/|A| \quad (13)$$

$$v(A) = d(A)/t(A) \quad (14)$$

Where $|A|$ denotes the area of the rectangle, $d(A)$ and $t(A)$ denotes the total distance and time travelled of all vehicles within this area. $N$ is the total number of vehicles, and $i$ represents the i-th vehicle. $\Delta x_i$, $\Delta t_i$ represent the total distance and time traveled by each vehicle. $q_{flow}, k_{density}, v(A)$ denote the traffic flow, density and space-mean speed.

## F. Reproducible trajectory extraction

All merging vehicles are filtered using lanelet2ID and latLaneCenterOffset. Nearby vehicles include eight vehicles with rearID, leadID, leftleadID, rightleadID, leftalongsideID, rightalongsideID, leftrearID, rightreadID. We match the nearest lead, rear and alongside vehicles on the outside of the mainline lane from position B to position F. Our code is under "/src/extraction/merging/main.py". The main loop of this code is reading tracks, recordingMeta, tracksMeta files of location 2, 3, 5, 6. The subloop contains: a) traversing each vehicle and determining whether the vehicle is a mainline, on-ramp, and off-ramp vehicle based on its laneletID list. b) extracting all the frames with lanechange=1, eliminate the frames with smaller interval IDs, and determine whether it is an on-ramp lane-changing behavior c) matching the longitudinal and lateral speed, acceleration and position of the rear and lead vehicle.

Our extraction code contains the following parameters: a) TIMESTEP: The default parameter value is 0.04 seconds, which represents the interval between two frames collections. b) LOOKBACK: The data of each LOOKBACK frame before lanechange=1 is chosen, and the default value is set to 5. We set this variable to guarantee the precision of the extracted trajectory. c) DISTANCE: The surrounding vehicles within this range are filtered according to this value. We perform a sensitivity analysis on the setting of this threshold to explore whether the change in its value will affect the scenario classification.

Fig. 5 exhibits eight random extracted trajectories. The latlanecenteroffset values jump from positive 2 to negative 2 in the frame LANECHANGE =1. This indicates that the center point of a vehicle moved from the original lane to the target lane. The vehicle in subfigure a, after merging into the outer lane of the mainline and then immediately changing lanes to the inner lane, experiences two jumps from 2 to -2 in the value of its latLaneCenterOffset. Tab. 2 in Appendix A in gives the statistics of vehicles crossing the solid line to merge directly into the mainline via area 1 under each location. It can be found that in locations 2 and 5, there are 96 and 82 cars crossing area 1 respectively, which account for 96/1392 and 82/1655 of all merging vehicles.

**Tab. 2 Count of vehicles merging from area 1 to the mainline**

| description | location | car | truck | van |
|---|---|---|---|---|
| Merging into mainline lane from area 1 | 2 | 96 | 2 | 2 |
| | 3 | 0 | 0 | 0 |
| | 5 | 82 | 0 | 0 |
| | 6 | 1 | 0 | 1 |

## IV. MERGING BEHAVIOR ANALYSIS RESULTS

### A. Overall merging point, speed and duration analysis

From Fig. 3, it can be found that cars merge into the mainline at all positions of the acceleration lane, while trucks and vans merge into the mainline more in the middle and rear part of the acceleration lane. The boxplots of speed, distance, distance ratio, duration, lateral maximum velocity and lateral maximum acceleration are given in Fig. 6. Outliers are filtered using the IQR (interquartile range) of the box plot. Any object with an IQR that is M times smaller than the first quartile or M times larger than the third quartile is considered an outlier. M indicates the percentage of exceedance and is a factor that can be adjusted according to the actual situation. We set this value to 3 in this paper. It can be found that: a) merging speed: the mean and median for cars are significantly higher than for vans and trucks. Trucks have a smaller mean and median value. The first and third quartile spacing for trucks and vans are much smaller than the corresponding spacing for cars. b) merging distance: trucks have significantly higher distances than cars and vans, while the average distance of cars is lower than those of others; c) merging duration: cars have a shorter preparation time. d) maximum lateral speed and acceleration: the lateral speed of cars is significantly higher than that of trucks and vans, and the first and third quartile spacing are significantly greater than those of others. e) outliers: cars have more extreme values than vans and trucks, which are more concentrated in lateral speed and acceleration.

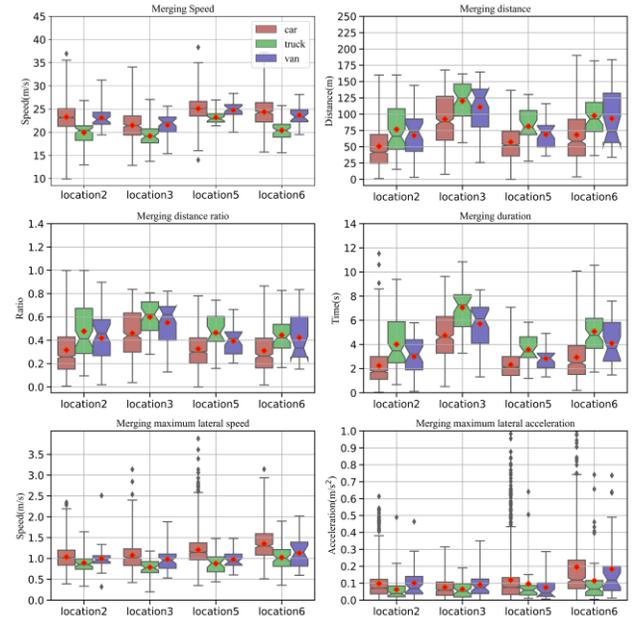

**Fig. 6 Boxplots of the merging characteristics of different types of vehicles at different locations**

### B. Clarification of the DISTANCE threshold and scenario division

Merging behaviors are classified into eight scenarios as shown in Tab. 3. "None" means this vehicle does not exist, "Exist" means that this vehicle exists, "Exist (rear to lead)" means that this previous rear vehicle becomes the new lead vehicle, and "Exist (lead to rear)" means that the previous lead vehicle becomes the new rear vehicle. If the lead or alongside vehicle changes to rear, the merging vehicle may speed up or other vehicles give way. According to vehicles types, locations, and eight scenarios, there are a total of $3\times3\times3\times4\times8=288$ subdivision scenarios. The determination of whether there is a lead or rear vehicle involves the determination of the DISTANCE threshold. When this value is exceeded, it is considered that there is no lead or rear vehicle. Existing studies have not divided it, nor has it given an accepted value to choose this value. We need to observe whether similar conclusions are reached under different threshold. Therefore, we choose three values for comparative analysis: 100m, 150m, and 200m.

**Tab. 3 Eight different types of merging scenarios**

| Scenario type | Rear vehicle | Lead vehicle |
|---|---|---|
| A | None | None |
| B | None | Exist |
| C | None | Exist (rear to lead) |
| D | Exist | None |
| E | Exist | Exist |
| F | Exist | Exist (rear to lead) |
| G | Exist (lead to rear) | None |
| H | Exist (lead to rear) | Exist |

We do not immediately apply various algorithms commonly used in existing studies, but rather divide the scenarios based on the change in the relative position of the vehicle with its surroundings. These algorithms need the determination of input variables, and the outcomes are impacted to some extent by the dataset or hyperparameters. This leads to the problem of insufficient detail in the trajectory extraction and vehicle matching, and the difficulty of reproducing the results of these studies by others. However, the foundational division method we employ in this study is applicable to any case and is unaffected by data, algorithms, or hyperparameters. A deeper level of merging pattern identification can be performed on the basis of our results. We argue that the scenarios need to be split as we done, rather than immediately adopting various algorithms that do not consider the DISTANCE threshold or the matching of surrounding vehicles.

## C. Accepted Gaps analysis

Tab. 5 in Appendix A exhibits the number of occurrences of different scenarios. Scenario G and H are excluded since only a single sample point is available. It can be found that: a) as the DISTANCE threshold becomes larger, the proportion of Scenario A, B, C and D gradually decreases, while the proportion of Scenario E and F gradually increases. This is due to the fact that there is originally no lead or rear vehicle in scenario A, B, C, D. As the threshold increases, more vehicles are discriminated as lead and rear vehicles. b) Scenario B and Scenario E appear the most times, both of which together could reach at least 65%, followed by scenario A and scenario D. Scenario C and scenario F have the least number of occurrences, with a total share of less than 6%. Fig. 7, Fig. 8, Fig. 15, Fig. 16, exhibit the boxplot of the distance headway and time headway with rear and lead vehicles.

It can be found that under different thresholds, the distance and time headway with the lead vehicle in scenarios C and F are significantly smaller than that in scenarios B and E, while the boxplot distributions for scenarios B and E are roughly similar. The average time headway with the lead vehicle in location 2 is about 2s, while this value is less than 1s in locations 3,5,6. This is in line with our common sense that when the original rear car becomes the final lead car, its spacing at the merging point is most likely to be at a lower level. Meanwhile, finding the ideal DISTANCE value seems to be another challenge for us, so we will investigate whether the variation of this value will affect our in-depth analysis.

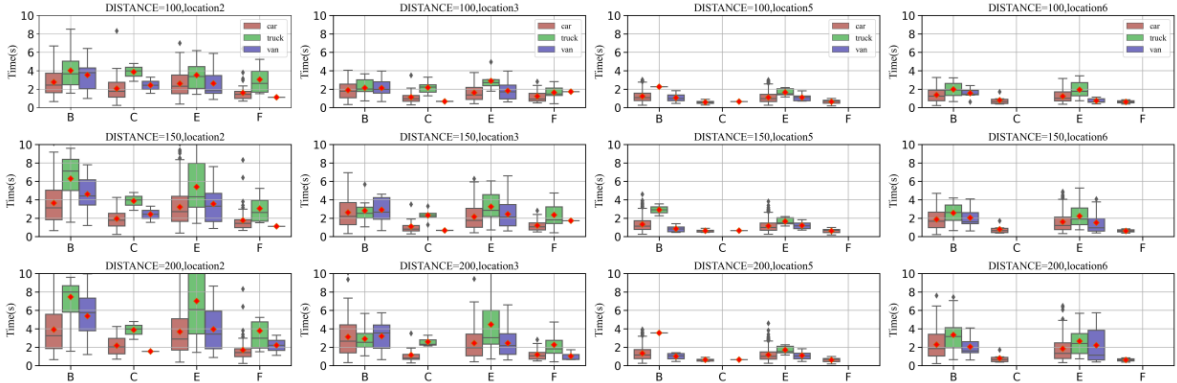

**Fig. 7 Boxplots of the time headway with the lead vehicles under different DISTANCE threshold**

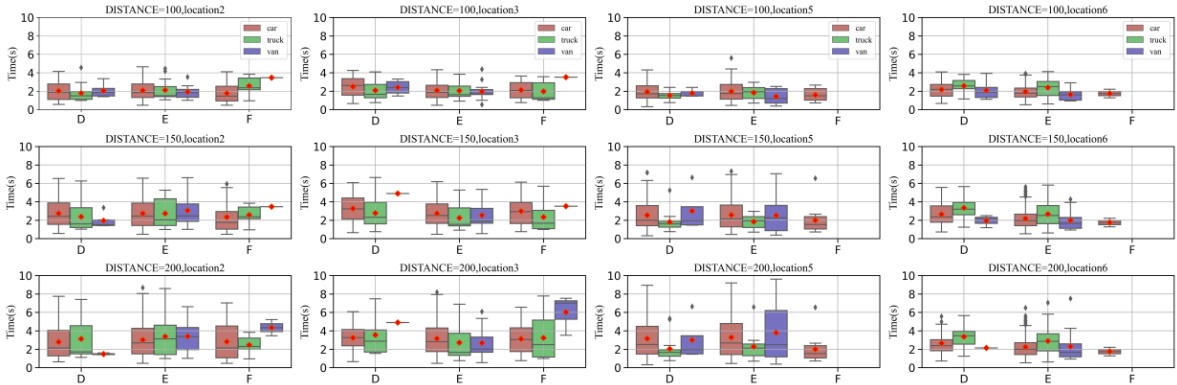

**Fig. 8 Boxplots of the time headway with the rear vehicles under different DISTANCE threshold**

## D. Similarity analysis

Jensen-Shannon (JS) divergence is introduced to measure the similarity between different scenarios. Given any two probability densities $f(x)$ and $g(x)$, JS divergence measures the correlation between these two probability densities. Compared with Kullback-Leibler (KL) divergence, JS divergence solves the problem that KL divergence is asymmetric. When two probability distributions are the same, the JS scatter is equal to 0, while when the probability distribution is more different, the JS value is closer to 1. Therefore, both the symmetry of JS divergence and the range of the value domain make it possible to discern the difference between two probability distributions more exactly.

$$D_{KL}(f \parallel g) = \int_{-\infty}^{\infty} f(x) \log(\frac{f(x)}{g(x)}) dx \quad (15)$$

$$D_{JS}(f \parallel g) = \frac{1}{2} D_{KL}(f \parallel \frac{f+g}{2}) + \frac{1}{2} D_{KL}(g \parallel \frac{f+g}{2}) \quad (16)$$

Where $D_{JS}(f \parallel g) = D_{JS}(g \parallel f)$. The range of JS values is between 0 and 1. Due to the small number of occurrences of truck and van in several scenarios, the distribution cannot be fitted, and the corresponding JS divergence cannot be obtained. Therefore, we only explore the JS divergence of cars. Fig. 9, Fig. 10, Fig. 17, Fig. 18 and Fig. 19 exhibit the heatmap of JS divergence values. The following findings could be found.

a) There was a large variability in the JS divergence values of these five variables among the four locations. For example, the JS divergence distribution of merging speed for different scenarios in location 2 differs significantly from those in location 5 or location 6. When the DISTANCE value is 100, the maximum JS values of merging speed corresponding to location 2 and location 6 are 0.30 and 1 respectively, the maximum JS values of distance ratio are 0.42 and 1 respectively, the maximum JS values of duration are 0.54 respectively and 1, the maximum JS values of lateral speed are 0.18 and 0.75 respectively, and the maximum JS values of lateral acceleration are 0.13 and 0.73 respectively. Although variations in DISTANCE thresholds also lead to differences in the distribution of JS divergence values, these differences are much smaller than the differences between locations. This suggests that there are large differences in merging behavior between locations, with smaller magnitudes of differences between merging scenarios in some locations (e.g., location 2 vs. location 3) and larger magnitudes of differences in others (e.g., location 5 vs. location 6). We could at least conclude that it is unreliable to analyze data from all locations together, and it is necessary to compare the variability between locations.

b) The JS divergence values of the distance ratio and duration are significantly higher than those of speed, maximum lateral speed and acceleration under different scenarios and locations (mean value is given in Tab. 6 and Tab. 7 in Appendix A). Maximum lateral speed and acceleration have the lowest JS divergence values, and the values of the merging speed is in the middle level. When the location is 2 and the DISTANCE threshold is 100m, the average JS divergence value of the distance ratio and duration is significantly higher than the other three variables, while the JS divergence values of lateral speed and acceleration are almost less than 0.1. This indicates that the three variables of merging speed, distance ratio and duration are more responsive to the differences between merging scenarios.

c) Since there are no rear cars in both Scenario A and Scenario B, the merging distance ratio and duration of these two scenarios are significantly lower than those in other scenarios either overall or at each location as exhibited in Tab. 6 and Tab. 7. When the DISTANCE threshold is 100m, the corresponding distance ratio are 0.38 and 0.36, and the duration are 3.48s and 3.29s as shown in Tab. 6 and Tab. 7. However, by observing the JS scatter values of scenario A and scenario B at different locations and threshold sizes and different variables, we found that the two scenarios are not always the closest. This indicates that although the mean values of the distance ratio and duration are closer for these two scenarios, there are differences between these two scenarios, which may be due to the presence of the lead car in scenario B.

d) Due to the presence of rear vehicles in both scenario D and scenario E, the average values of merging distance ratio and duration corresponding to these two scenarios are higher than those of scenario A and scenario B, but lower than those of scenario C and scenario F as shown in Tab. 6 and Tab. 7. When the DISTANCE threshold is 100m, the values of distance ratio and duration of these scenarios are 0.5/0.46 and 4.45s/4.17s. Compared to scenario A and scenario B, there is a significant difference in the distance ratio and duration for the presence or absence of a rear vehicle, both at the overall level, at the location level, and at the level of each vehicle type. However, the JS scatter values under each variable for these two scenarios are not always the closest, probably because there is also a front car in scenario E. This leads to the fact that the two merging scenarios are still different and cannot be grouped together.

e) The heatmaps of JS divergence among different variables at different locations both show that scenario C and scenario F are significantly different from other scenarios with divergence values much higher than the mean. Although the overall frequency of these two scenarios is below 5%, its distance ratio and duration are higher than those of the other scenarios as shown in Tab. 6 and Tab. 7. This is largely due to the fact that in both scenarios, the vehicle in front of the merging vehicle at the moment of the merging point was originally the vehicle behind the merging vehicle. There is a parallel driving situation between the merging vehicle and this vehicle. These two vehicles are more likely to have a high probability of strong interaction in the lateral as well as the longitudinal direction. The speed of the vehicle may be higher than that of the merging vehicle, or the merging vehicle may slow down to give way. Therefore, the merging behavior of vehicles may have been affected, causing them to choose longer merge distances and times. The other four types of scenarios do not involve alongside vehicles, so we believe that the merging behavior in scenarios C and F is more worthy of our subsequent in-depth analysis. It is worth noting that these two types of scenarios have not been extracted in existing research, and they often group into scenario E or scenario B.

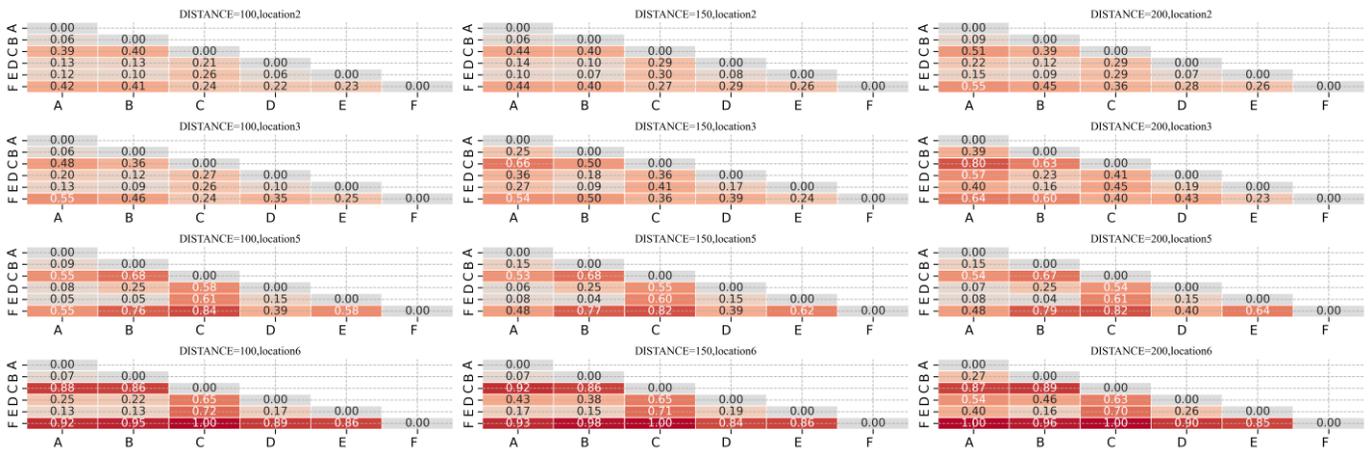

**Fig. 9 Heatmap of JS divergence of merging distance ratio under different DISTANCE threshold**

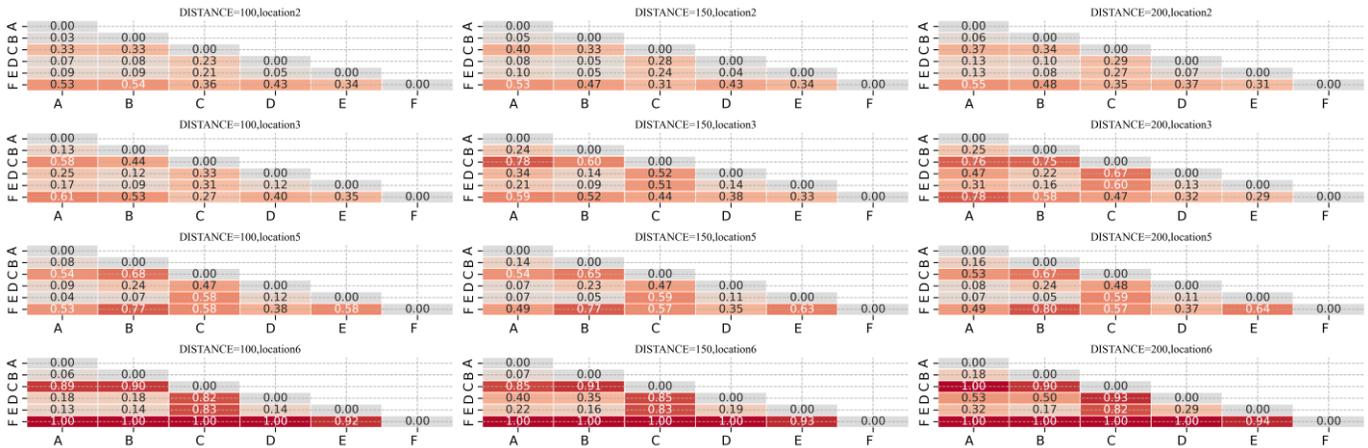

**Fig. 10 Heatmap of JS divergence of merging duration under different DISTANCE threshold**

*E. Safety analysis*

Fig. 20 in Appendix A shows the boxplot of TTC values with the lead and rear vehicle. It can be found that scenarios C and F have significantly lower TTC values under each DISTANCE threshold, under different types of vehicles, and under each location than those of other scenarios. Since trucks and vans do not have samples in some scenarios, we only give the mean histogram of TTC for cars under different DISTANCE thresholds at different locations (as shown in Fig. 11). The average TTC values of scenario B in locations 2, 3, and 5 are higher than those of other scenarios, while the average TTC of scenario B/ D/ E in location 6 are closer. Among all locations, the TTC values of scenario D and scenario E are closer. The average TTC values of scenario C are lower than 2s in locations 2,3,6 and lower than 3s in location 5. In locations 2,3,6, the average TTC value of scenario F is slightly higher than that of scenario C. However, in scenario 5, the TTC value of scenario F is significantly higher than that of scenario C.

*F. Consecutive lane-changing duration analysis*

Fig. 21 in Appendix A exhibits the occurrence percentages and durations of merging vehicles moving from the outside lane to the inside lane. Tab. 8 and Tab. 9 in Appendix A gives the specific values. The average values of different DISTANCE thresholds are given on the far right of Fig. 21. It can be found that the number of occurrences and duration of this behavior on different locations has a large difference. Scenarios B and E always have the highest number of occurrences, while scenarios C, D, and F have fewer occurrences. Scenarios C and F in location 5 have a proportion of 0, while scenario D in locations 2,3,6 have the smallest proportion of occurrences with values of 0.08, 0.21, 0.16, respectively.

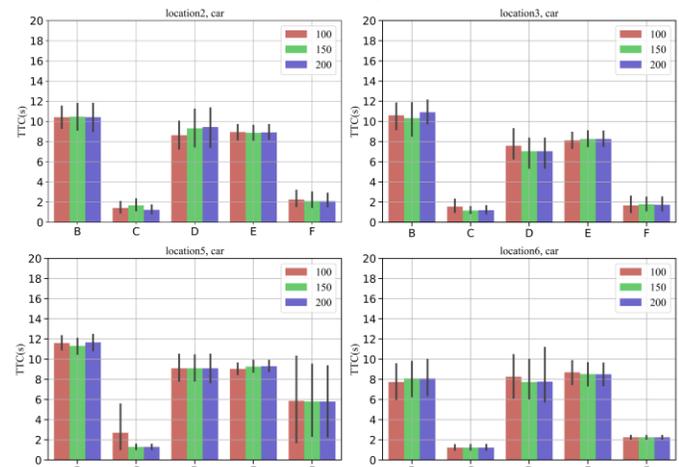

**Fig. 11 TTC values for different merging scenarios**

*G. Traffic flow-density-speed analysis*

The information at the macroscopic traffic flow level is more or less already included in the classified scenarios, mainly due to the setting of DISTANCE threshold. The merging behavior in scenario A does not have a lead or rear vehicle, and its corresponding traffic flow may be at a low level. The merging vehicles in scenarios C, B, and D all have a vehicle lead or a rear vehicle, and the corresponding traffic flow may be at an intermediate level. The merging vehicles in scenario E and F all have a lead and rear vehicles, and the corresponding traffic flow may be at a high level, especially since the merging vehicle in scenario F is inserted into the gap that was originally the two cars behind it. The calculation time range of the flow, density, speed we choose is from position B to position F, and the spatial range is the upstream and downstream of the merging area in Fig. 4. Within our reproducible framework, interested readers can tune these parameters. Fig. 13 shows the traffic flow, density and speed, and Fig. 12 exhibits the average value under different scenarios.

It could be found that: a) Scenarios A, B, and C do not have rear vehicles, so their corresponding upstream traffic flows are lower than those of other scenarios, with the average values of 128.40 veh/h, 153.02 veh/h, and 339.33 veh/h for the four locations, respectively. Scenarios D, E, and F have rear vehicles, so their upstream traffic flows are at a higher level, with the average values of 495.40veh/h, 557.05veh/h, and 537.36veh/h for the four locations, respectively. b) There is no vehicle in front in scenario A and scenario D, so the corresponding downstream traffic flow is lower than those of other scenarios. The average value of the four locations is 428veh/h and 520.35veh/h. The traffic flows of scenarios B, C, E, and F are at relatively high levels, and the average flows of the four locations are 854.77veh/h, 866.66 veh/h, 949.70 veh/h, and 1084.20 veh/h, respectively. c) The upstream traffic flow curves in locations 2, 3, and 5 show a gradual increase, and the scenario with the highest traffic flow in these three locations is scenario F, while the scenario with the highest traffic flow in location 6 is scenario E. The downstream traffic flow curves for the four locations show an increase, then a decrease, and then an increase again. The maximum downstream traffic flow for locations 2, 3, and 6 occurs in scenario F, while the maximum for location 5 occurs in scenario C.

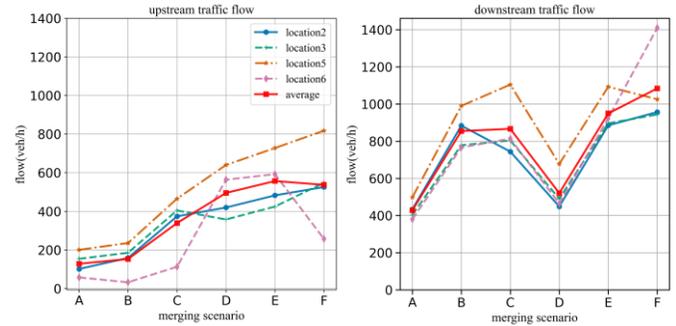

**Fig. 12 upstream and downstream traffic flow**

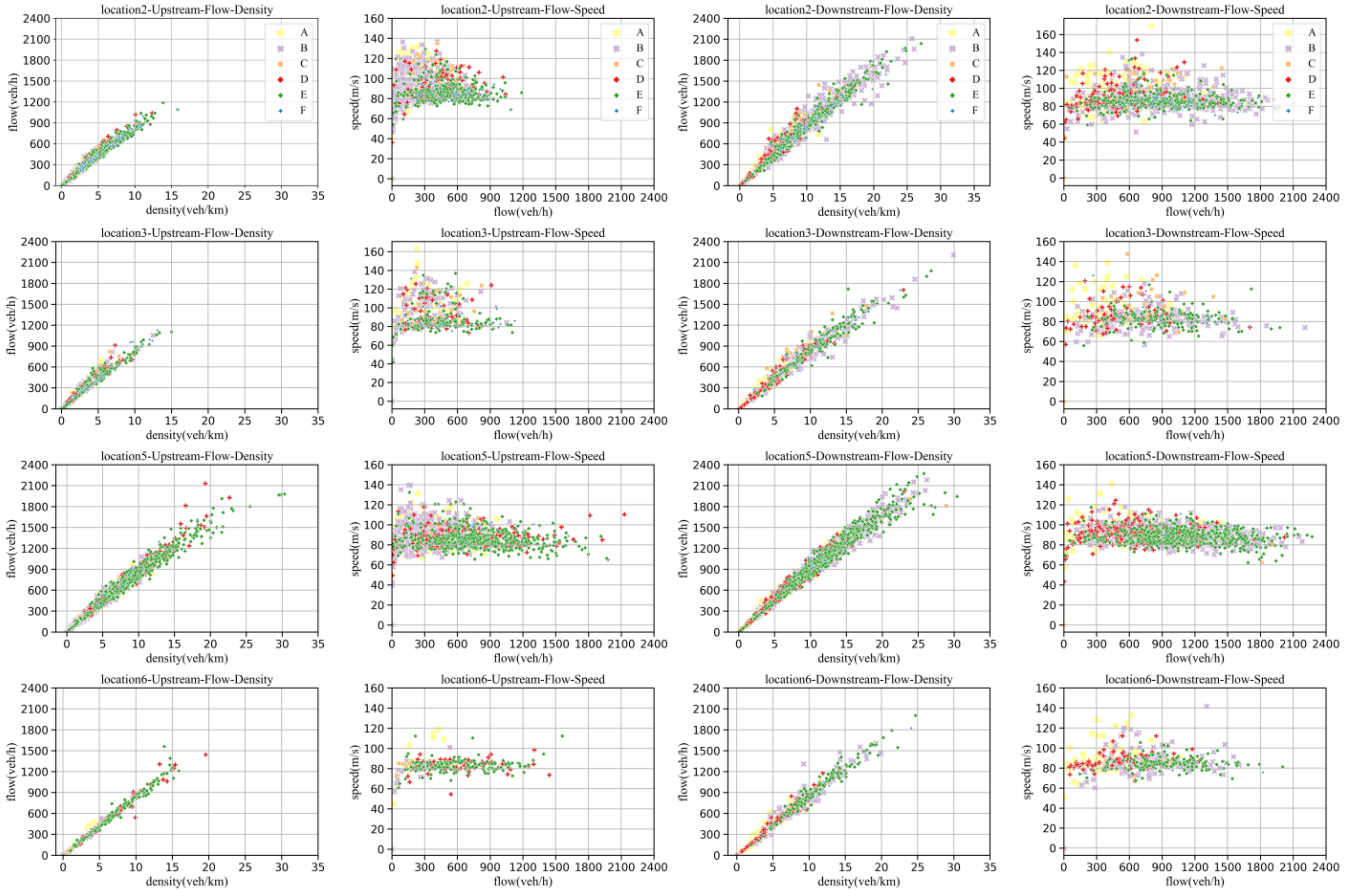

**Fig. 13 Scatter diagram of traffic flow, density and speed upstream and downstream of the merging area**

## V. MAKING RESEARCH RESULTS REPRODUCIBLE

The reproducible architecture of this study is not limited to making this paper reproducible, but also to guide everyone's driving behavior analysis work to be reproducible by others.

### A. Why reproducibility is needed?

Our results should be not only be verified by oneself, but also by others. Recent years have seen an unprecedented increase in interest in reproducible research. The main benefits of reproducible research for oneself, for the reader, and for society includes: 1) improved work and work habits; 2) improved teamwork; 3) greater impact: less inadvertent competition and more acknowledgement; 4) greater continuity and cumulative impact; 5) stewardship of public goods; 6) public access to public goods [20, 30].

### B. Why this research is more reproducible?

This paper is more easily to be reproduced by others. First, the causes of subjectivity and irreproducibility in existing studies are minimized. Second, we define the standardized indicators and clarifies the time-space range. Third, we present a transparent analysis, with all source code and intermediate process being made public. Finally, we form a Github project and provide reproduction instructions.

### C. How to reproduce this paper?

We provide all the source codes associated to any of the charts (as displayed in Tab. 4). This will enhance the reliability of our discoveries and make it easier for readers to utilize this work as a preliminary guide. The reproduction steps of our analysis results can be referred to Fig. 14, including: a) Download the project, set up the environment, label the HD map data, and configure the settings. b) Start the main.py code, which is used for trajectory extraction and indicators calculation. c) All the figures and tables can be reproduced by running each python file under the figure and table folder.

### D. Inspiration for porting this research to other locations

Many existing comparative analyses of merging locations (different regions or countries) are often subjective and non-reproducible, which makes the results less convincing. The main reason for this is the difficulty of standardizing the analysis process. Just by analyzing the trajectory dataset, it is difficult to unify various criteria, such as the identification of position B, D and F in Fig. 4. If we employ HD map of the merging area for comparative analysis of different locations, we could unify various standards, such as trajectory segmentation and extraction, indicators definition and calculation, alongside vehicle identification and scenario classification. Following the reproducible framework of this paper, a standardized analysis of merging behavior can effectively avoid the subjectivity that exists in groups, space and time, and thus better highlight new findings.

### E. Inspiration for porting this research to other behaviors

The reproducible approach of this study is not only limited to merging behavior, but also applicable to other driving behaviors, such as lane-changing behavior, cut-in behavior, car-following behavior, etc. For example, during the lane-changing or cut-in behavior, the methods of segmenting the spatial-temporal position of each trajectory based on HD map in this study is also applicable to identifying the moment when the center of the lane-changing vehicle falls on the road lane. Through the latLaneCenterOffset field, the lateral trajectory, speed and acceleration could be investigated. Meanwhile, it is also possible to explore whether there are alongside vehicles.

**Tab. 4 Source code files for each figure and table in this paper**

| Chart | Input data | Caption | Reproduction address |
|---|---|---|---|
| Figures | asset/mergingData100m.csv<br>asset/mergingData150m.csv<br>asset/mergingData200m.csv | Figure 3 | src/figure/preliminaryAnalysis/mergingPointScatterToMap.py |
| | | Figure 5 | src/figure/preliminaryAnalysis/randomNineMergingTrajectory.py |
| | | Figure 6 | src/figure/preliminaryAnalysis/boxplotMergingPointDistanceDuration.py |
| | | Figure 7 | src/figure/acceptedGapsAnalysis/LeadTimeHeadway.py |
| | | Figure 8 | src/figure/acceptedGapsAnalysis/RearTimeHeadway.py |
| | | Figure 9 | src/figure/similarityAnalysis/JSDivergence.py |
| | | Figure 10 | |
| | | Figure 17 | |
| | | Figure 18 | |
| | | Figure 19 | |
| | | Figure 11 | src/figure/safetyAnalysis/carSafetyAnalysis.py |
| | | Figure 12 | src/figure/trafficFlowSpeedDensity/trafficFlowMergingType.py |
| | | Figure 13 | src/figure/trafficFlowSpeedDensity/trafficSpeedFlowScatter.py |
| | | Figure 15 | src/figure/acceptedGapsAnalysis/LeadDistanceHeadway.py |
| | | Figure 16 | src/figure/acceptedGapsAnalysis/RearDistanceHeadway.py |
| | | Figure 20 | src/figure/safetyAnalysis/boxplotofTTC.py |
| | | Figure 21 | src/figure/consecutiveLanechangeAnalysis/occurrenceAndDuration.py |
| Tables | | Table 1 | asset/table/laneletIdAndLength.xlsx |
| | | Table 2 | asset/table/countMergingFromArea1.csv |
| | | Table 5 | src/table/EightMergingTypeStatistic.py |
| | | Table 6 | |
| | | Table 7 | |
| | | Table 8 | src/table/consecutiveLaneChangingDuration.py |
| | | Table 9 | |

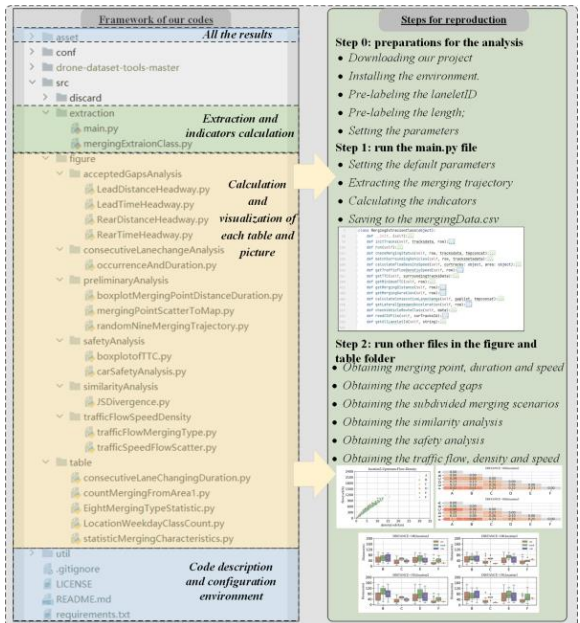

**Fig. 14 Project structure and steps for reproduction**

## VI. FINDINGS AND DISCUSSION

### A. Importance of this topic

*1) The need for more fined-grained analysis.*

The trajectory extraction, vehicle matching, indicator definition, scenario classification in existing studies are not refined enough. The incorporation of HD map allows us to capture the spatial-temporal trajectory in a more refined manner, define each indicator more objectively, perform vehicle match and scenario classification more precisely.

*2) The need for reproducible research*

Existing studies undoubtedly contributed to our understanding of merging behavior, but they might be accompanied by a reproducible crisis. We may need to minimize the existence of such kind of crisis. Only in this manner will the results of various application of algorithm be more reliable and likely to propel research in this field forward.

### B. Summary of novel findings

We construct distance ratio indicator to enable comparisons across horizontally (various scenarios at the same location) and vertically (different locations). Our results suggest that this indicator could effectively reflect the preference of vehicles for the choice of the merging point. We identify vehicles that cross the solid line to merge into the main line as exhibited in Tab. 2. Local traffic management could optimize the design of the marking of the merging area to further reduce the safety risk using this indicator. Under different DISTANCE thresholds, we found that the occurrence percentage of scenarios C and F is much lower than others, but their behavior characteristics are significantly different from other scenarios as reflected in the heatmap of JS divergence. Meanwhile, the merging distance, duration and risk of these two scenarios are significantly higher than those of other scenarios. This suggests that the vehicle in these scenarios prepares for a considerable period of time before driving into the main lane and these two scenarios are more likely to be high-risk. We also found that there is no situation where the JS divergence value corresponding to two scenarios is always minimum or maximum for different DISTANCE thresholds, different variables, and different locations. For example, the JS values between scenario A and B, and between scenarios D and E are not always the lowest. This suggests that different scenarios delineated all have their own characteristics. These scenarios cannot simply be grouped arbitrarily in a crude manner as in existing studies.

In summary, our novel findings can be distilled down to: a) Site-specific analysis is essential, as the characteristics of each location may be different. b) Each merging scenario has its own characteristics and cannot be easily combined for analysis. c) The distance ratio and the duration variables better reflect the differences in merging behavior. d) The merging behavior of the presence of alongside vehicles is significantly different from other scenarios and is often accompanied by a higher risk of travel.

### C. The application value of this paper

The abnormal data points in each variable identified using box plots are more likely to be corner cases in real-world driving environments. Although existing simulation tests methods can achieve certain results by fully parameterizing the scenarios, these scenarios generated are unrealistic due to their overly random parameter selection. This leads to low efficiency in scenario construction and simulation testing. This paper could guide the construction of merging testing scenarios, parameter sampling, and the setting of the proportion of different subdivision scenarios, especially in scenario C and scenario F. Meanwhile, anyone can conduct more sophisticated mining based on reproducible frameworks. For instance, it is vitally important to go further for scenarios C and F, including investigating whether there are any significant interactions between merging vehicles and their surroundings, how to quantify these interactions, and how to include them in the decision-planning algorithm

### D. Long-term effects and potential benefits

The long-term effect of this study includes: presenting a unique approach to merging behavior analysis, directing readers in reproducing any part of this paper, and even digging farther based on our framework. The potential benefits of this study lie in the promotion of reproducible research concept. This will not only be helpful in promoting our understanding of merging behavior, but also in extending it to car-following behavior, lane-changing behavior, etc.

### E. Limitations and future work

Providing the characteristics of each driver while capturing tens of thousands of vehicle trajectories by drones seems unlikely to be possible for a long time to come. Although natural driving data provide information about drivers, it is difficult to provide large-scale trajectory datasets and HD map data. Additionally, the data obtained by means of simulation cannot reflect real-world driving scenario. Therefore, it is important to strike a balance between dataset size, driver privacy, data authenticity, and collection accuracy. Whether it is to dig deeper into the analysis, or establish some data-driven models, or transplant our analysis framework to location 0 and location 1, or combine high-precision maps to analyze other driving behaviors, can be fully based on our framework.

## VII. CONCLUSION

HD map is employed to analyze the merging behavior in a fine-grained way for the first time. Our results demonstrate that the application of HD map does contribute to our deeper understanding of the merging behavior, and provides a direction to address the shortcomings of the existing studies. The proposed reproducible approach is not only applicable to this research, but also for other merging locations and even other driving behaviors. We argue that only when the merging data processing and analysis are sufficiently refined, it will be more likely to advance the relevant data-driven research. It is hoped that this study would raise awareness of the significance of the reproducible research.

## VIII. APPENDIX A

To keep the main body of the article concise, we have placed some figures and tables in this section. Our code project is at https://github.com/iyangli/A-Reproducible-approach-to-merging-behavior-analysis-based-on-High-Definition-Map.git, which consists of:

*1) Directory "asset"*
  a) Extracted merging trajectory (under three DISTANCE);
  b) All figures;
  c) All tables.

*2) Directory "conf"*
  a) lanelet2 HD map data;
  b) visualization parameters;
  c) laneletID pre-labeling file. laneletID.py contains the division of each area as shown in Fig. 4, and the length of each laneletID.

*3) Directory "src"*
  a) subdirectory extraction includes trajectory extraction and metrics calculation, file export;
  b) subdirectory acceptedGapsAnalysis provides the source code for accepted gap analysis, including Fig. 7, Fig. 8, Fig. 15, and Fig. 16;
  c) subdirectory consecutiveLanechangeAnalysis provides the source code for consecutive lane-changing duration analysis, including Fig. 21, Tab. 8, Tab. 9;
  d) subdirectory preliminaryAnalysis provides the source codes for preliminary analysis of merging behavior, including Fig. 5, Fig. 6, and Fig. 3;
  e) subdirectory safetyAnalysis provides the source code for safety analysis, including Fig. 11 and Fig. 20;
  f) subdirectory similarityAnalysis provides the source codes for JS value calculation, including Fig. 9, Fig. 10, Fig. 17, Fig. 18, Fig. 19;
  g) subdirectory trafficFlowSpeedDensity provides the source codes for traffic flow-density-speed analysis, including Fig. 12 and Fig. 13

*4) Directory "util"*
  Public functions when processing data.

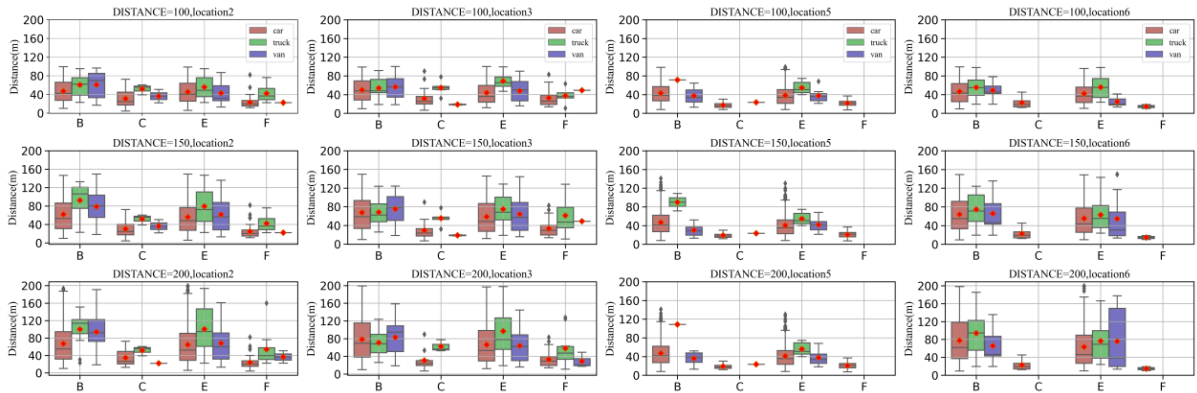

**Fig. 15 Boxplots of the distance headway with the lead vehicles under different DISTANCE threshold**

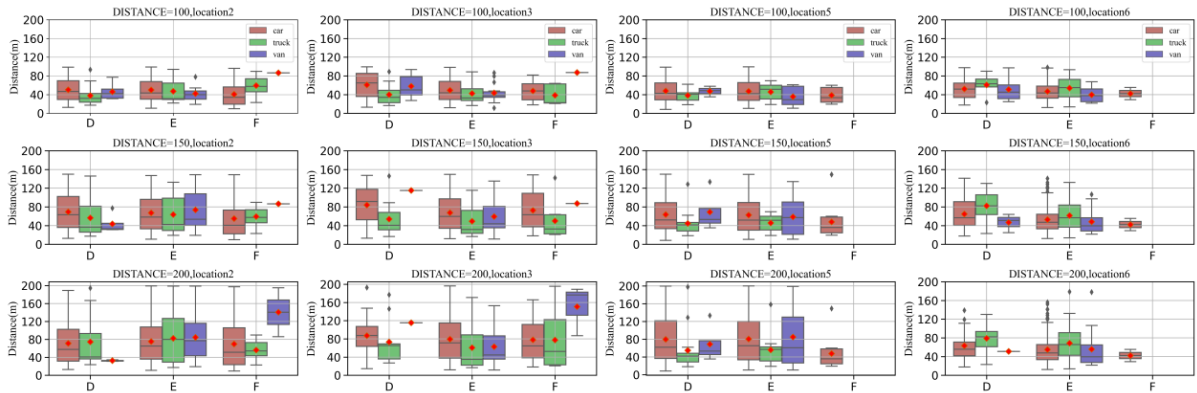

**Fig. 16 Boxplots of the distance headway with the rear vehicles under different DISTANCE threshold**

**Tab. 5 Counts for different types of merging scenarios under different DISTANCE threshold**

| location | vehicle class | DISTANCE THRESHOLD = 100m | | | | | | DISTANCE THRESHOLD = 150m | | | | | | DISTANCE THRESHOLD = 200m | | | | | |
|---|---|---|---|---|---|---|---|---|---|---|---|---|---|---|---|---|---|---|---|
| | | A | B | C | D | E | F | A | B | C | D | E | F | A | B | C | D | E | F |
| 2 | total | 254 | 451 | 45 | 142 | 346 | 54 | 105 | 442 | 36 | 124 | 522 | 63 | 68 | 423 | 27 | 89 | 612 | 73 |
| | car | 222 | 425 | 40 | 119 | 327 | 44 | 89 | 415 | 31 | 106 | 483 | 53 | 59 | 398 | 23 | 77 | 559 | 61 |
| | truck | 26 | 11 | 3 | 17 | 11 | 9 | 14 | 14 | 3 | 14 | 23 | 9 | 9 | 12 | 3 | 9 | 34 | 10 |
| | van | 6 | 15 | 2 | 6 | 8 | 1 | 2 | 13 | 2 | 4 | 16 | 1 | | 13 | 1 | 3 | 19 | 2 |
| 3 | total | 101 | 175 | 30 | 79 | 153 | 27 | 36 | 167 | 19 | 56 | 247 | 40 | 18 | 153 | 13 | 39 | 296 | 46 |
| | car | 88 | 154 | 22 | 58 | 131 | 21 | 30 | 147 | 12 | 43 | 211 | 31 | 14 | 135 | 10 | 29 | 253 | 33 |
| | truck | 6 | 9 | 6 | 16 | 7 | 5 | 4 | 8 | 5 | 12 | 12 | 8 | 3 | 6 | 3 | 9 | 18 | 10 |
| | van | 7 | 12 | 2 | 5 | 15 | 1 | 2 | 12 | 2 | 1 | 24 | 1 | 1 | 12 | | 1 | 25 | 3 |
| 5 | total | 209 | 599 | 7 | 203 | 542 | 11 | 143 | 481 | 6 | 235 | 694 | 12 | 106 | 350 | 6 | 272 | 825 | 12 |
| | car | 202 | 586 | 6 | 187 | 522 | 11 | 139 | 471 | 5 | 217 | 670 | 12 | 103 | 345 | 5 | 253 | 796 | 12 |
| | truck | 4 | 1 | | 13 | 10 | | 2 | 2 | | 14 | 10 | | 1 | 1 | | 15 | 11 | |
| | van | 3 | 12 | 1 | 3 | 10 | | 2 | 8 | 1 | 4 | 14 | | 2 | 4 | 1 | 4 | 18 | |
| 6 | total | 112 | 167 | 6 | 71 | 157 | 2 | 54 | 198 | 6 | 51 | 204 | 2 | 23 | 225 | 6 | 34 | 225 | 2 |
| | car | 90 | 147 | 6 | 55 | 127 | 2 | 44 | 175 | 6 | 36 | 164 | 2 | 18 | 199 | 6 | 25 | 177 | 2 |
| | truck | 17 | 13 | | 10 | 21 | | 7 | 15 | | 12 | 27 | | 3 | 18 | | 8 | 32 | |
| | van | 5 | 7 | | 6 | 9 | | 3 | 8 | | 3 | 13 | | 2 | 8 | | 1 | 16 | |
| total | | 676 | 1392 | 88 | 495 | 1198 | 94 | 338 | 1288 | 67 | 466 | 1667 | 117 | 215 | 1151 | 52 | 434 | 1958 | 133 |
| Percentage | | 17.14% | 35.30% | 2.23% | 12.55% | 30.38% | 2.38% | 8.57% | 32.67% | 1.70% | 11.82% | 42.28% | 2.97% | 5.45% | 29.19% | 1.32% | 11.01% | 49.66% | 3.37% |

**Tab. 6 Mean distance ratio under different merging scenarios under different DISTANCE threshold**

| location | vehicle class | DISTANCE THRESHOLD = 100m | | | | | | DISTANCE THRESHOLD = 150m | | | | | | DISTANCE THRESHOLD = 200m | | | | | |
|---|---|---|---|---|---|---|---|---|---|---|---|---|---|---|---|---|---|---|---|
| | | A | B | C | D | E | F | A | B | C | D | E | F | A | B | C | D | E | F |
| 2 | total | 0.32 | 0.32 | 0.52 | 0.50 | 0.44 | 0.73 | 0.28 | 0.33 | 0.52 | 0.48 | 0.40 | 0.73 | 0.27 | 0.30 | 0.49 | 0.53 | 0.41 | 0.68 |
| | car | 0.27 | 0.24 | 0.52 | 0.37 | 0.37 | 0.58 | 0.25 | 0.25 | 0.50 | 0.34 | 0.34 | 0.58 | 0.23 | 0.23 | 0.44 | 0.35 | 0.35 | 0.59 |
| | truck | 0.38 | 0.38 | 0.55 | 0.51 | 0.53 | 0.73 | 0.39 | 0.43 | 0.55 | 0.55 | 0.41 | 0.73 | 0.32 | 0.41 | 0.55 | 0.70 | 0.40 | 0.73 |
| | van | 0.30 | 0.34 | 0.51 | 0.61 | 0.43 | 0.90 | 0.19 | 0.32 | 0.51 | 0.56 | 0.45 | 0.90 | | 0.26 | 0.48 | 0.55 | 0.47 | 0.72 |
| 3 | total | 0.48 | 0.45 | 0.67 | 0.53 | 0.58 | 0.67 | 0.49 | 0.45 | 0.66 | 0.60 | 0.54 | 0.67 | 0.50 | 0.43 | 0.60 | 0.60 | 0.55 | 0.69 |
| | car | 0.39 | 0.41 | 0.62 | 0.49 | 0.50 | 0.65 | 0.38 | 0.39 | 0.62 | 0.48 | 0.48 | 0.64 | 0.36 | 0.35 | 0.64 | 0.50 | 0.49 | 0.63 |
| | truck | 0.56 | 0.46 | 0.66 | 0.64 | 0.62 | 0.69 | 0.59 | 0.51 | 0.63 | 0.60 | 0.57 | 0.72 | 0.53 | 0.52 | 0.55 | 0.60 | 0.57 | 0.73 |
| | van | 0.49 | 0.49 | 0.74 | 0.48 | 0.62 | 0.66 | 0.49 | 0.45 | 0.74 | 0.71 | 0.58 | 0.66 | 0.60 | 0.42 | | 0.71 | 0.59 | 0.71 |
| 5 | total | 0.36 | 0.35 | 0.43 | 0.48 | 0.38 | 0.57 | 0.36 | 0.31 | 0.41 | 0.48 | 0.37 | 0.57 | 0.32 | 0.25 | 0.41 | 0.47 | 0.37 | 0.57 |
| | car | 0.35 | 0.26 | 0.46 | 0.47 | 0.33 | 0.57 | 0.39 | 0.26 | 0.43 | 0.46 | 0.31 | 0.57 | 0.38 | 0.25 | 0.43 | 0.45 | 0.31 | 0.57 |
| | truck | 0.31 | 0.44 | | 0.57 | 0.39 | | 0.28 | 0.33 | | 0.56 | 0.39 | | 0.17 | 0.22 | | 0.55 | 0.39 | |
| | van | 0.40 | 0.35 | 0.39 | 0.41 | 0.43 | | 0.40 | 0.34 | 0.39 | 0.41 | 0.41 | | 0.40 | 0.27 | 0.39 | 0.41 | 0.42 | |
| 6 | total | 0.36 | 0.32 | 0.66 | 0.47 | 0.42 | 0.65 | 0.34 | 0.31 | 0.66 | 0.47 | 0.44 | 0.65 | 0.26 | 0.31 | 0.66 | 0.53 | 0.45 | 0.65 |
| | car | 0.26 | 0.24 | 0.66 | 0.39 | 0.37 | 0.65 | 0.25 | 0.23 | 0.66 | 0.44 | 0.37 | 0.65 | 0.23 | 0.23 | 0.66 | 0.46 | 0.37 | 0.65 |
| | truck | 0.50 | 0.39 | | 0.40 | 0.45 | | 0.42 | 0.43 | | 0.46 | 0.45 | | 0.36 | 0.42 | | 0.49 | 0.45 | |
| | van | 0.30 | 0.31 | | 0.62 | 0.45 | | 0.36 | 0.28 | | 0.52 | 0.51 | | 0.20 | 0.28 | | 0.62 | 0.51 | |
| Total mean | | 0.38 | 0.36 | **0.57** | 0.50 | 0.46 | **0.68** | 0.37 | 0.35 | **0.56** | 0.51 | 0.44 | **0.68** | 0.34 | 0.32 | **0.52** | 0.53 | 0.44 | **0.67** |

**Tab. 7 Mean duration under different merging scenarios under different DISTANCE threshold**

| location | vehicle class | DISTANCE THRESHOLD = 100m | | | | | | DISTANCE THRESHOLD = 150m | | | | | | DISTANCE THRESHOLD = 200m | | | | | |
|---|---|---|---|---|---|---|---|---|---|---|---|---|---|---|---|---|---|---|---|
| | | A | B | C | D | E | F | A | B | C | D | E | F | A | B | C | D | E | F |
| 2 | total | 2.43 | 2.38 | 4.51 | 3.70 | 3.28 | 5.48 | 2.19 | 2.53 | 4.40 | 3.60 | 2.95 | 5.51 | 2.19 | 2.33 | 4.21 | 4.05 | 3.03 | 5.34 |
| | car | 1.85 | 1.68 | 4.19 | 2.49 | 2.56 | 4.86 | 1.75 | 1.73 | 3.85 | 2.29 | 2.36 | 4.94 | 1.65 | 1.61 | 3.46 | 2.36 | 2.39 | 4.95 |
| | truck | 3.33 | 3.11 | 5.13 | 4.35 | 4.00 | 6.10 | 3.57 | 3.55 | 5.13 | 4.54 | 3.28 | 6.10 | 2.72 | 3.60 | 5.13 | 5.80 | 3.30 | 6.13 |
| | van | 2.12 | 2.36 | 4.22 | 4.25 | 3.28 | 5.48 | 1.26 | 2.30 | 4.22 | 3.96 | 3.22 | 5.48 | | 1.79 | 4.04 | 3.99 | 3.40 | 4.94 |
| 3 | total | 5.37 | 4.75 | 7.08 | 5.96 | 6.22 | 7.12 | 5.36 | 4.88 | 7.17 | 6.64 | 5.83 | 6.98 | 5.49 | 4.76 | 6.55 | 6.59 | 5.95 | 7.31 |
| | car | 4.10 | 4.24 | 6.39 | 4.97 | 5.04 | 7.10 | 3.95 | 4.07 | 6.54 | 5.08 | 4.86 | 6.81 | 3.55 | 3.75 | 6.77 | 5.20 | 4.96 | 6.72 |
| | truck | 6.74 | 5.04 | 7.24 | 7.91 | 7.22 | 7.94 | 6.84 | 5.86 | 7.36 | 7.72 | 6.66 | 7.81 | 6.29 | 6.00 | 6.33 | 7.46 | 6.93 | 8.03 |
| | van | 5.26 | 4.98 | 7.62 | 5.00 | 6.41 | 6.32 | 5.30 | 4.71 | 7.62 | 7.12 | 5.97 | 6.32 | 6.64 | 4.54 | | 7.12 | 5.98 | 7.19 |
| 5 | total | 2.52 | 2.63 | 3.37 | 3.53 | 2.84 | 4.37 | 2.47 | 2.30 | 3.23 | 3.50 | 2.74 | 4.45 | 2.19 | 1.83 | 3.23 | 3.45 | 2.74 | 4.45 |
| | car | 2.40 | 1.85 | 3.98 | 3.21 | 2.43 | 4.37 | 2.64 | 1.82 | 3.70 | 3.17 | 2.28 | 4.45 | 2.58 | 1.79 | 3.70 | 3.12 | 2.22 | 4.45 |
| | truck | 2.38 | 3.52 | | 4.44 | 2.92 | | 2.12 | 2.64 | | 4.37 | 2.92 | | 1.32 | 1.76 | | 4.27 | 2.98 | |
| | van | 2.77 | 2.51 | 2.76 | 2.96 | 3.18 | | 2.66 | 2.43 | 2.76 | 2.97 | 3.03 | | 2.66 | 1.93 | 2.76 | 2.97 | 3.01 | |
| 6 | total | 3.63 | 3.38 | 7.52 | 4.63 | 4.32 | 8.38 | 3.51 | 3.31 | 7.52 | 4.76 | 4.50 | 8.38 | 2.81 | 3.31 | 7.52 | 5.23 | 4.51 | 8.38 |
| | car | 2.44 | 2.28 | 7.52 | 3.57 | 3.47 | 8.38 | 2.30 | 2.12 | 7.52 | 4.01 | 3.51 | 8.38 | 2.22 | 2.11 | 7.52 | 4.25 | 3.54 | 8.38 |
| | truck | 5.74 | 4.68 | | 4.54 | 5.03 | | 5.09 | 4.96 | | 4.99 | 5.17 | | 4.45 | 4.97 | | 5.40 | 5.11 | |
| | van | 2.70 | 3.19 | | 5.77 | 4.46 | | 3.15 | 2.85 | | 5.29 | 4.81 | | 1.76 | 2.85 | | 6.04 | 4.89 | |
| Total mean | | 3.48 | 3.29 | **5.45** | 4.45 | 4.17 | **6.32** | 3.39 | 3.25 | **5.41** | 4.63 | 4.01 | **6.29** | 3.26 | 3.06 | **4.96** | 4.83 | 4.06 | **6.35** |

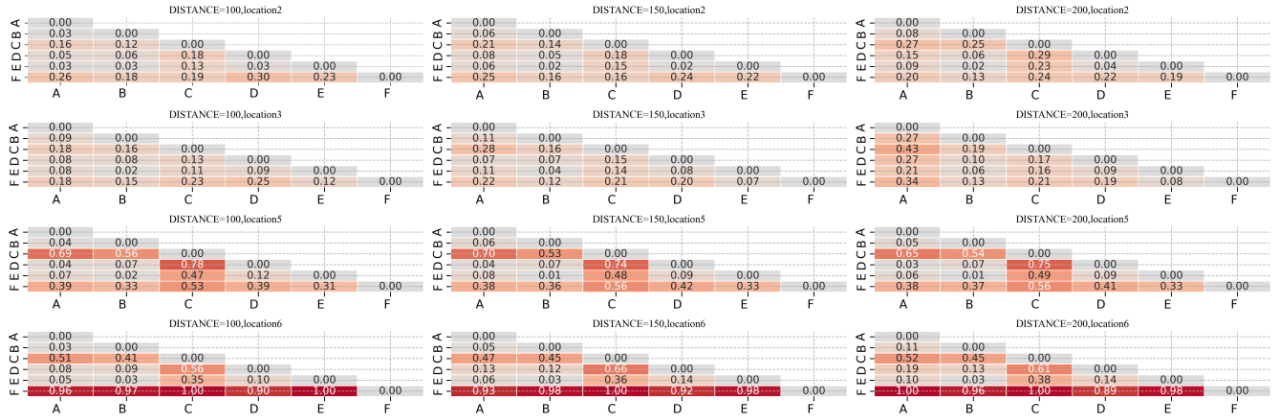

**Fig. 17 Heatmap of JS divergence of merging speed under different DISTANCE threshold**

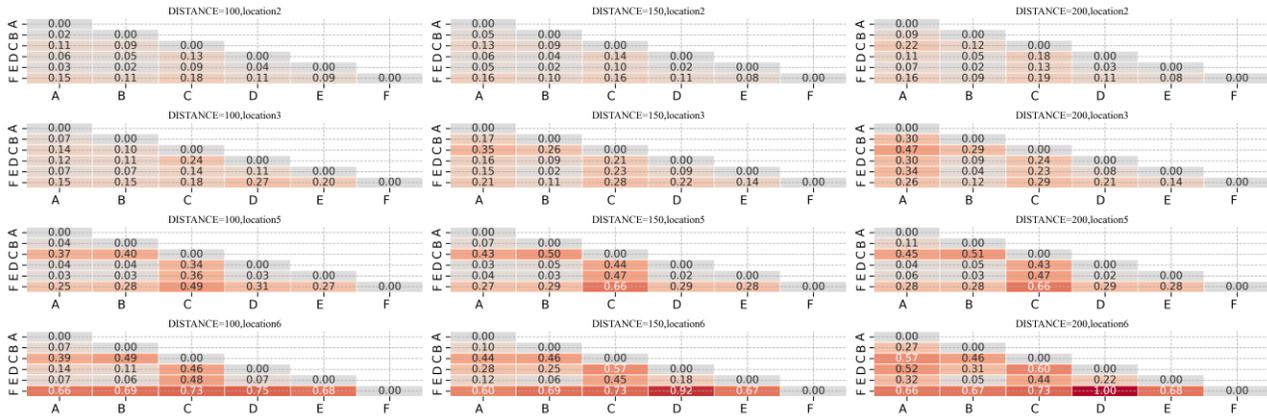

**Fig. 18** Heatmap of JS divergence of maximum lateral speed under different DISTANCE threshold

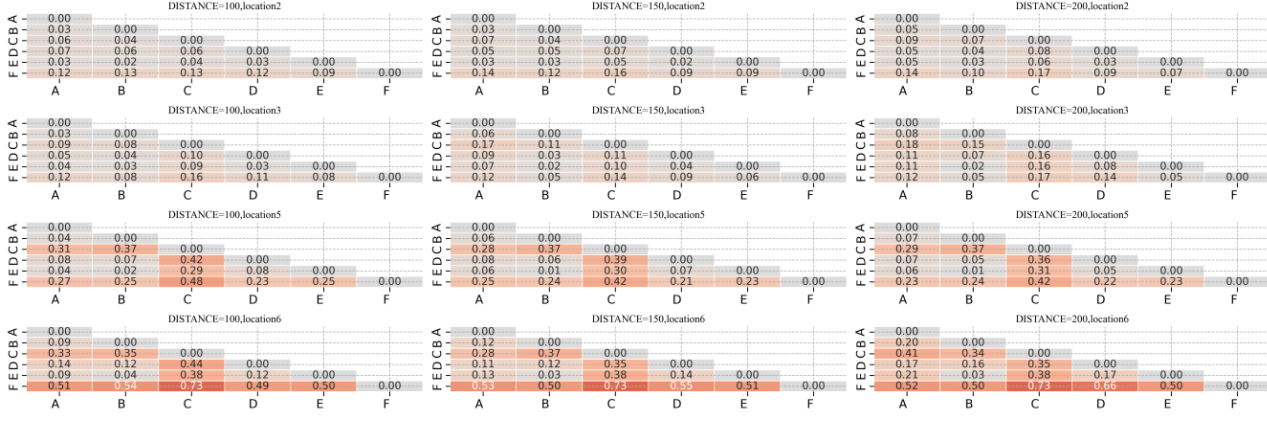

**Fig. 19** Heatmap of JS divergence of maximum lateral acceleration under different DISTANCE threshold

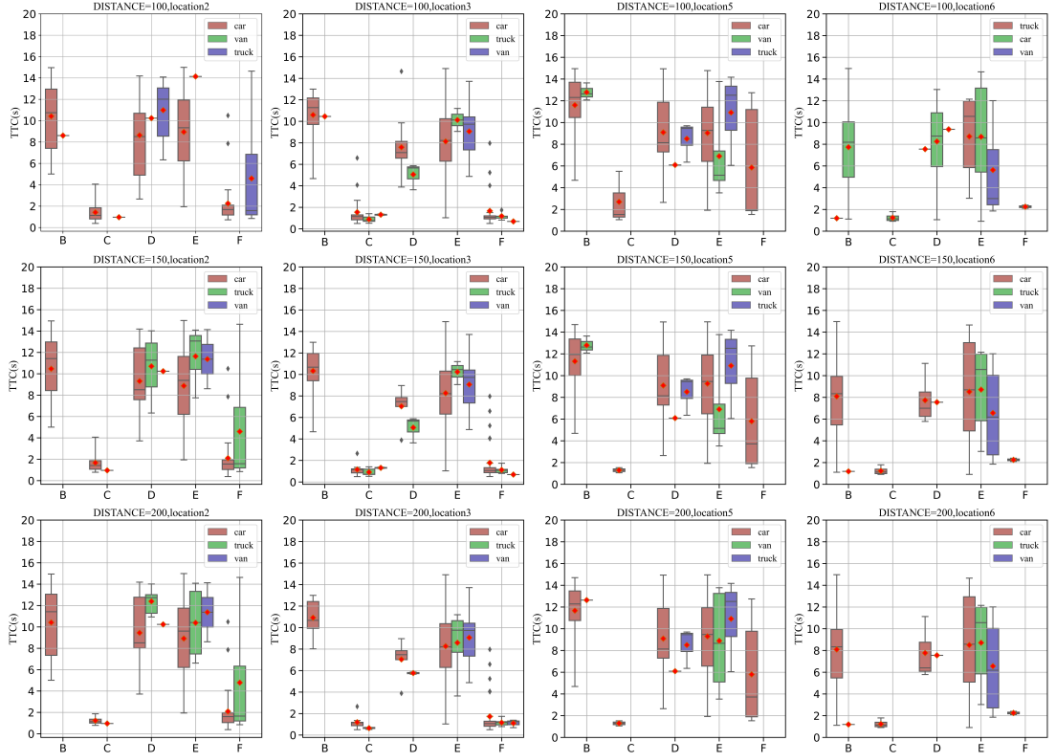

**Fig. 20** Boxplots of the minimum TTC with rear and lead vehicle under different DISTANCE threshold

**Tab. 8** Percentage of two consecutive lane changes occurring under different DISTANCE threshold

| location | vehicle class | DISTANCE THRESHOLD = 100m | DISTANCE THRESHOLD = 150m | DISTANCE THRESHOLD = 200m |
| --- | --- | --- | --- | --- |

|  |  | A | B | C | D | E | F | A | B | C | D | E | F | A | B | C | D | E | F |
|---|---|---|---|---|---|---|---|---|---|---|---|---|---|---|---|---|---|---|---|
| 2 | total | 0.04 | 0.24 | 0.06 | 0.03 | 0.18 | 0.39 | 0.03 | 0.18 | 0.06 | 0.03 | 0.18 | 0.39 | 0.06 | 0.18 | 0.06 | 0.02 | 0.16 | 0.23 |
|  | car | 0.11 | 0.35 | 0.18 | 0.09 | 0.34 | 0.18 | 0.10 | 0.30 | 0.19 | 0.08 | 0.31 | 0.17 | 0.12 | 0.30 | 0.17 | 0.06 | 0.29 | 0.18 |
|  | truck |  | 0.09 |  |  | 0.09 |  |  | 0.07 |  |  | 0.04 |  |  | 0.08 |  |  | 0.03 |  |
|  | van |  | 0.27 |  |  | 0.13 | 1.00 |  | 0.15 |  |  | 0.19 | 1.00 |  | 0.15 |  |  | 0.16 | 0.50 |
| 3 | total | 0.30 | 0.27 | 0.08 | 0.16 | 0.21 | 0.13 | 0.55 | 0.27 | 0.08 | 0.06 | 0.20 | 0.11 | 0.61 | 0.31 | 0.10 | 0.06 | 0.19 | 0.11 |
|  | car | 0.32 | 0.39 | 0.23 | 0.28 | 0.37 | 0.38 | 0.40 | 0.40 | 0.25 | 0.19 | 0.35 | 0.32 | 0.50 | 0.43 | 0.20 | 0.17 | 0.33 | 0.33 |
|  | truck | 0.17 |  |  |  |  |  | 0.25 |  |  |  | 0.33 |  |  |  |  |  |  |  |
|  | van | 0.43 | 0.42 |  | 0.20 | 0.27 | 1.00 | 0.42 |  |  |  | 0.25 |  | 1.00 | 0.50 |  |  | 0.24 |  |
| 5 | total | 0.03 | 0.09 |  | 0.01 | 0.09 |  | 0.02 | 0.09 |  | 0.01 | 0.08 |  | 0.02 | 0.10 |  | 0.01 | 0.08 |  |
|  | car | 0.09 | 0.26 |  | 0.04 | 0.16 |  | 0.06 | 0.27 |  | 0.04 | 0.18 |  | 0.05 | 0.30 |  | 0.04 | 0.18 |  |
|  | truck |  |  |  |  |  |  |  |  |  |  |  |  |  |  |  |  |  |  |
|  | van |  |  |  |  | 0.10 |  |  |  |  |  | 0.07 |  |  |  |  |  | 0.06 |  |
| 6 | total | 0.07 | 0.29 | 0.17 | 0.08 | 0.45 |  | 0.11 | 0.25 | 0.17 | 0.03 | 0.36 |  | 0.09 | 0.24 | 0.17 | 0.01 | 0.32 |  |
|  | car | 0.22 | 0.64 | 0.17 | 0.25 | 0.62 |  | 0.32 | 0.55 | 0.17 | 0.08 | 0.57 |  | 0.28 | 0.53 | 0.17 | 0.04 | 0.54 |  |
|  | truck |  | 0.08 |  |  | 0.05 |  |  | 0.07 |  |  | 0.04 |  |  | 0.06 |  |  | 0.03 |  |
|  | van |  | 0.14 |  |  | 0.67 |  |  | 0.13 |  |  | 0.46 |  |  | 0.13 |  |  | 0.38 |  |
| Total mean | | 0.38 | 0.11 | 0.22 | 0.06 | 0.07 | 0.23 | 0.20 | 0.18 | 0.20 | 0.07 | 0.03 | 0.21 | 0.19 | 0.21 | 0.21 | 0.07 | 0.03 | 0.19 |

**Tab. 9 Two consecutive lane-changing duration under different DISTANCE threshold**

| location | vehicle class | DISTANCE THRESHOLD = 100m | | | | | | DISTANCE THRESHOLD = 150m | | | | | | DISTANCE THRESHOLD = 200m | | | | | |
|---|---|---|---|---|---|---|---|---|---|---|---|---|---|---|---|---|---|---|---|
|  |  | A | B | C | D | E | F | A | B | C | D | E | F | A | B | C | D | E | F |
| 2 | total | 5.18 | 4.96 | 5.86 | 5.75 | 4.12 | 5.35 | 4.27 | 4.78 | 5.81 | 6.29 | 4.42 | 5.30 | 4.17 | 4.79 | 6.69 | 7.16 | 4.41 | 5.03 |
|  | car | 5.18 | 4.88 | 5.86 | 5.75 | 5.09 | 7.10 | 4.27 | 4.93 | 5.81 | 6.29 | 5.05 | 6.99 | 4.17 | 4.96 | 6.69 | 7.16 | 5.02 | 6.46 |
|  | truck |  | 4.92 |  |  | 3.00 |  |  | 4.92 |  |  | 3.00 |  |  | 4.92 |  |  | 3.00 |  |
|  | van |  | 5.07 |  |  | 4.28 | 3.60 |  | 4.48 |  |  | 5.20 | 3.60 |  | 4.48 |  |  | 5.20 | 3.60 |
| 3 | total | 7.35 | 5.13 | 5.56 | 3.82 | 4.68 | 5.81 | 7.23 | 5.47 | 4.93 | 3.13 | 4.50 | 5.95 | 7.25 | 5.48 | 5.94 | 2.78 | 4.49 | 5.67 |
|  | car | 3.33 | 4.92 | 5.56 | 4.12 | 4.61 | 5.81 | 3.46 | 4.44 | 4.93 | 3.13 | 4.76 | 5.95 | 4.06 | 4.23 | 5.94 | 2.78 | 4.73 | 5.67 |
|  | truck | 10.84 |  |  |  |  |  | 10.84 |  |  |  |  |  | 10.84 |  |  |  |  |  |
|  | van | 7.88 | 5.34 |  | 3.52 | 4.74 |  | 7.38 | 6.50 |  |  | 4.25 |  | 6.84 | 6.74 |  |  | 4.25 |  |
| 5 | total | 2.41 | 2.69 |  | 2.86 | 3.05 |  | 2.36 | 2.64 |  | 2.97 | 3.06 |  | 2.26 | 2.61 |  | 2.84 | 3.06 |  |
|  | car | 2.41 | 2.69 |  | 2.86 | 2.71 |  | 2.36 | 2.64 |  | 2.97 | 2.73 |  | 2.26 | 2.61 |  | 2.84 | 2.73 |  |
|  | truck |  |  |  |  |  |  |  |  |  |  |  |  |  |  |  |  |  |  |
|  | van |  |  |  |  | 3.40 |  |  |  |  |  | 3.40 |  |  |  |  |  | 3.40 |  |
| 6 | total | 5.27 | 5.81 | 7.92 | 3.77 | 6.15 |  | 5.73 | 5.81 | 7.92 | 2.91 | 6.08 |  | 2.57 | 5.89 | 7.92 | 4.16 | 6.06 |  |
|  | car | 5.27 | 4.70 | 7.92 | 3.77 | 5.30 |  | 5.73 | 4.71 | 7.92 | 2.91 | 5.09 |  | 2.57 | 4.94 | 7.92 | 4.16 | 5.03 |  |
|  | truck |  | 7.44 |  |  | 8.04 |  |  | 7.44 |  |  | 8.04 |  |  | 7.44 |  |  | 8.04 |  |
|  | van |  | 5.28 |  |  | 5.11 |  |  | 5.28 |  |  | 5.11 |  |  | 5.28 |  |  | 5.11 |  |
| Total mean | | 5.82 | 5.03 | 6.45 | 4.01 | 4.63 | 5.50 | 5.67 | 5.04 | 6.22 | 3.83 | 4.66 | 5.51 | 5.12 | 5.07 | 6.85 | 4.23 | 4.65 | 5.24 |

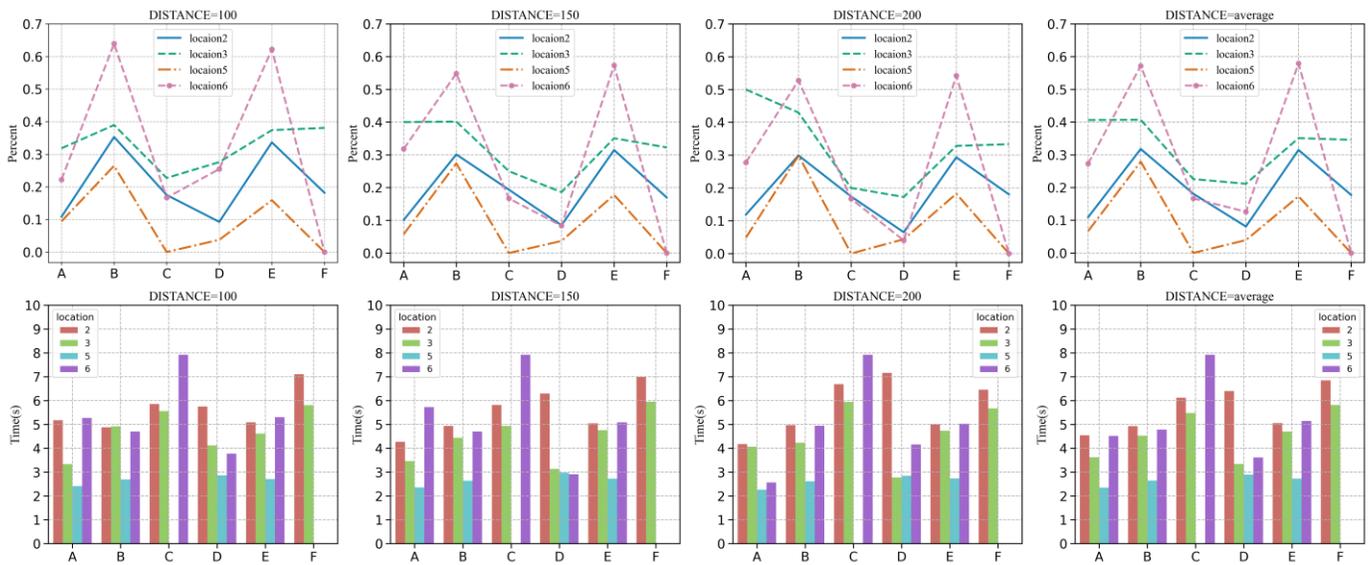

**Fig. 21 The occurrence percentages and durations of merging vehicles moving from the outside lane to the inside lane**

## IX. ACKNOWLEDGMENT

The authors would like to thank the Editor and anonymous reviewers for their insightful and constructive comments that helped them improve this work greatly.

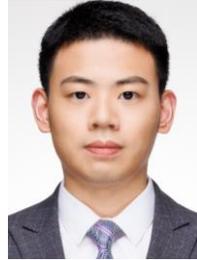

**Yang Li** was born in Shanghai, China. He is currently pursuing the Ph.D. degree with Tongji University. From June 2022 to the end of September 2022, he is doing an internship and research in the AI group of Huawei's autonomous driving solution' planning and controlling department. His main interests include traffic flow theory and simulation, connected vehicle technology, and traffic data analysis.

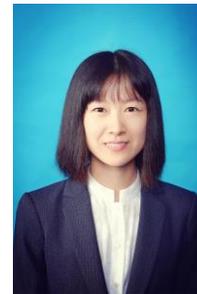

Dr. Yang Liu is Assistant Professor in the Department of Civil and Environmental Engineering, and the Department of Industrial Systems Engineering and Management at the National University of Singapore. She earned her B.S. from Tsinghua University, China, her MPhil from Hong Kong University of Science and Technology, Hong Kong, and her Ph.D. from Northwestern University, IL, USA. Her research focuses on future mobility and transport, covering topics such as ridesharing and carsharing systems operations and design, travel demand and congestion management, and data-driven transportation system modeling and analysis.

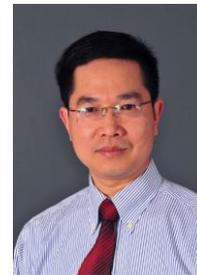

**Daiheng Ni** is Professor of Civil and Environmental Engineering at University of Massachusetts Amherst, USA. He received the Ph.D. degree in Civil Engineering from Georgia Institute of Technology, GA, in 2004. His research interests include traffic flow theory and simulation, connected vehicle technology, and Intelligent Transportation Systems. He is the author of two books and more than 50 refereed journal articles. He is an Associate Editor of Journal of Intelligent Transportation Systems, Taylor & Francis.

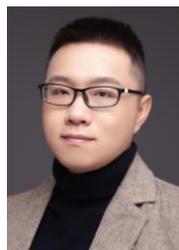

**Ang Ji** received the Ph.D. Degree in Engineering from University of Sydney, Australia, in 2022. He is currently an Assistant Professor at School of Transportation and Logistics, Southwest Jiaotong University. His research interests include connected and automated vehicles, microscopic traffic modeling, deep reinforcement learning applications, and game theoretical approach in traffic control.

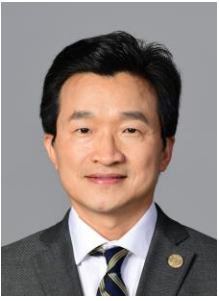

**Linbo Li** is Associate Professor of Transportation Engineering at Tongji University, CN. He received the Ph.D. degree in Transportation Engineering from Tongji University in 2007. His research interests include traffic flow theory, traffic planning and management, and smart parking. He is the author of two books and more than 30 refereed journal articles and was granted 6 national invention patents.

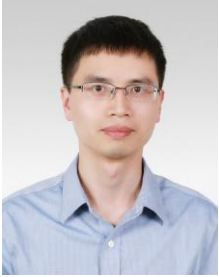

**Yajie Zou** received the B.S. degree in engineering mechanics from Tongji University, Shanghai, China, and the M.S. and Ph.D. degrees in transportation engineering from Texas A&M University. He is cur-rently an Associate Professor at Tongji University. His main research interests include traffic operations, traffic safety, and microscopic traffic simulation models.